\documentclass[10pt,twocolumn,letterpaper]{article}

\usepackage{cvpr}
\usepackage{times}
\usepackage{epsfig}
\usepackage{graphicx}
\usepackage{amsmath}
\usepackage{amssymb}
\usepackage{upgreek}
\usepackage{tabularx}  
\usepackage{float}
\usepackage[colorlinks,
linkcolor=blue,
anchorcolor=blue,
citecolor=blue
]{hyperref}
\usepackage{authblk}
\usepackage{booktabs}
\usepackage{multirow}

\cvprfinalcopy 

\begin{document}

\title{VPC-Net: Completion of 3D Vehicles from MLS Point Clouds}

\author[a]{Yan Xia}
\author[a]{Yusheng Xu}
\author[b]{Cheng Wang}
\author[a]{Uwe Stilla}
\affil[a]{Photogrammetry and Remote Sensing, Technical University of Munich (TUM) \authorcr \emph {\{yan.xia, yusheng.xu, stilla\}@tum.de} }
\affil[b]{Fujian Key Laboratory of Sensing and Computing, School of Informatics, Xiamen University \authorcr \emph{cwang@xmu.edu.cn} }

\maketitle

\begin{abstract}
\vspace{-0.25cm}

As a dynamic and essential component in the road environment of urban scenarios, vehicles are the most popular investigation targets. To monitor their behavior and extract their geometric characteristics, an accurate and instant measurement of vehicles plays a vital role in   traffic and transportation fields. 
Point clouds acquired from the mobile laser scanning (MLS) system deliver 3D information of road scenes with   unprecedented detail. 
They have proven to be an adequate data source in the fields of intelligent transportation and autonomous driving, especially for extracting vehicles. However, acquired 3D point clouds of vehicles from MLS systems are inevitably incomplete due to object occlusion or self-occlusion. To tackle this problem, we proposed a neural network to synthesize complete, dense, and uniform point clouds for vehicles from MLS data, named Vehicle Points Completion-Net (VPC-Net). In this network, we introduce a new encoder module to extract global features from the input instance, consisting of a spatial transformer network and point feature enhancement layer. Moreover, a new refiner module is also presented to preserve the vehicle details from inputs and refine the complete outputs with fine-grained information. Given sparse and partial point clouds as inputs, the network can generate complete and realistic vehicle structures  and keep the fine-grained details from the partial inputs. We evaluated the proposed VPC-Net in different experiments using synthetic and real-scan datasets and applied the results to 3D vehicle monitoring tasks. Quantitative and qualitative experiments demonstrate the promising performance of the proposed VPC-Net and show state-of-the-art results. 
\end{abstract}
\begin{figure*}[!tb]
	\begin{center}
		\includegraphics[width=16cm]{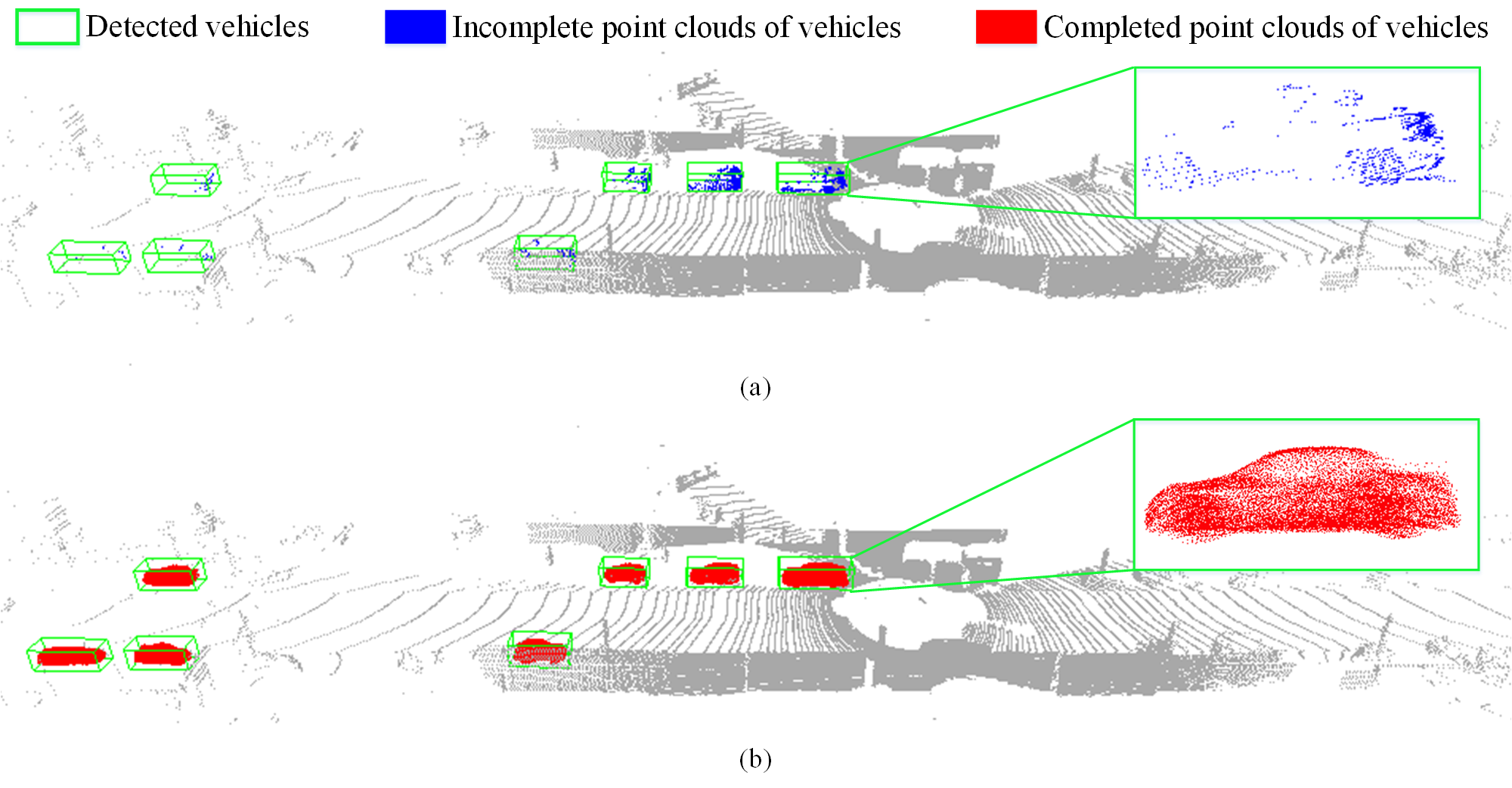}
		\setlength{\abovecaptionskip}{3pt}
		\setlength{\belowcaptionskip}{-12pt}		
		\caption{Illustration of the incomplete and completed point clouds of vehicles. (a) A single-frame raw real-scan data from KITTI \cite{geiger2013vision}. (b) Completed scan generated by the proposed Vehicle Points Completion-Net (VPC-Net).}
		\label{fig:cover}
    \end{center}
\end{figure*}
\section{Introduction}
In the road environment of urban scenarios, vehicles are the most popular investigation targets and   a dynamic and essential component.
For monitoring their behaviors and delineating their geometric characteristics, an accurate and instant measurement of vehicles plays a vital role. 
Measurements can be achieved via either cameras or laser scanners, providing 2D images \cite{tuermer2013airborne} and 3D point clouds \cite{yao2011extraction, yao2012airborne}, respectively. 
Compared with 2D images, 3D point clouds acquired from mobile laser scanning (MLS) systems deliver detailed 3D information of road scenes while 
   driving with a high measuring frequency. 
Acquiring point clouds via an MLS system provides an efficient solution for tasks such as autonomous driving and urban street mapping. Especially for vehicle extraction, MLS systems have been chosen as a key sensor by plenty of autonomous driving companies and research institutes, since it can provide highly accurate geometric information (e.g., 3D coordinates of vehicle points) and reliable ratiometric attributes (e.g., reflectivities of various surface materials) of multiple instances simultaneously.

However, acquired 3D point clouds of vehicles from MLS systems are inevitably incomplete due to object occlusion or self-occlusion ~\cite{peters2020conditional}. 
For instance, in Fig. \ref{fig:cover}a, a few typical point clouds of vehicles on urban roads from the KITTI dataset \cite{geiger2013vision} are illustrated. 
As seen from this figure, we can clearly observe the missing parts in vehicles' scanned point clouds. 
This incompleteness of measured data significantly hinders the potential uses of vehicle point clouds, because it actually has changed the dimension of shapes, biased the volume of objects, and destroyed the topology of surfaces. 
In generic applications such as 3D traffic monitoring, complete geometric shapes of vehicles provide solid foundations for 3D perceptual tasks, including instance extraction, type classification, and track estimation \cite{wen2019toward}.  
For example, in some specific applications such as the measure of vehicle-induced aerodynamic loads in bridge engineering,  the complete surface, as well as the shape, of the measured vehicles is the key to the estimation of 
  wind pressure caused by  vehicles driving close to the sound barrier, which has a considerable impact on designing the structure of urban highway viaducts \cite{pan2018novel}. 
However, it is almost impossible to directly obtain  complete point clouds of vehicles because the viewpoints of an MLS system is always fixed in the center of roads and can hardly observe vehicles from the facade direction. Moreover, for applications such as autonomous driving, the requirement of time efficiency makes it impossible to conduct multiple measurements to compensate for occlusions.
To eventually obtain complete point clouds of vehicles, many approaches using various strategies have been posed, for instance,    an alternative strategy based on 
a given database of CAD models \cite{zhang2020vehicle}, which first estimates the vehicle poses and then retrieves a similar CAD model of that vehicle from the large-scale CAD model datasets to replace the raw point clouds. However, this method cannot deal with occluded vehicles, and it is unable to preserve the real knowledge of raw point clouds. 
Apart from the database-based solutions, directly completing sparse and partial point clouds is a more practical and straightforward solution and can benefit from the popular machine learning methods with the help of sufficient training samples. 
Nevertheless, this is not an easy task since the incomplete point clouds are missing geometric and semantic information. The challenges are regarding the following aspects:

\begin{itemize}
    {\item how to guarantee an even distribution when generating a dense and complete point cloud;}
    {\item how to reconstruct the proper topology of missing shapes and keep the  fine-grained details of partial point clouds.}
\end{itemize}

In this paper, we propose a neural network, named Vehicle Points Completion-Net (VPC-Net), to synthesize complete, dense, and uniform point clouds for vehicles from MLS data. Given the sparse and partial point clouds of vehicles, our network can generate complete and realistic structures and keep the fine-grained details from the partial inputs, as shown in Fig.~\ref{fig:cover}b. The significant contributions are as follows:
\begin{itemize}
    \item We design a novel end-to-end network (termed as VPC-Net) for completing point clouds of 3D vehicle shapes, operating on the partial and sparse point clouds directly. By endorsing an architecture with the encoder, decoder, and refiner, VPC-Net can produce uniform, dense, and complete point clouds from partially scanned vehicles in MLS datasets.
    \item We propose a novel encoder module to better extract global features from the instance, including a spatial transformer network (STN) and a point feature enhancement (PFE) layer. The STN ensures the extracted features are invariant to geometric transformations from input point clouds with different resolutions. The PFE layer combines low-level and high-level information to enhance the feature representation ability.
    \item We propose a new refiner module to preserve the vehicle details from inputs and refine the complete outputs with fine-grained details. To fully retain the details of the input point cloud,  the partial inputs and outputs generated by the decoder are combined uniformly. A point feature residual network is designed to predict per-wise offsets for every point.
    \item We conduct experiments on one 3D synthetic dataset (i.e., ShapeNet \cite{chang2015shapenet}) and two real MLS datasets (i.e., KITTI \cite{geiger2013vision} and TUM-MLS-2016 \cite{zhu2020tum}), demonstrating that the proposed network VPC-Net achieves superior performance over the baseline methods. We adopt this method to 3D vehicle monitoring, which completes dynamic 3D vehicles of the 3D scene online only based on a single frame of raw real-scan data, not relying on the information of time sequences.
\end{itemize}

The remainder of this paper is organized as follows. Section~\ref{sec: related work} briefly reviews and discusses the related works for point cloud and shape completion. Section~\ref{sec: overview} introduces an overview of the proposed   VPC-Net. Section~\ref{sec: network} elaborates on the detailed architecture of the networks, including explanations on each module. Section~\ref{sec:experiments} presents the experiments. Experimental results are shown in Section~\ref{sec:Experimental results}, and Section~\ref{sec:Discussion} gives a detailed discussion and analysis of  the derived results.  Section~\ref{sec:Conclusions} concludes the paper and outlines future work.

\section{Related work}\label{sec: related work}

3D shape completion has long been an attractive research topic in robotics and computer vision for many years \cite{anguelov2005scape, han2017high}. There is now a series of methods for recovering complete geometric information from partial point clouds. Generally, the related methods can be primarily classified into three major categories: (i) geometry-based methods, (ii) template-based methods, and (iii) learning-based methods. In the following subsection, we briefly review these three types of methods.

\subsection{Geometry-based shape completion}

The geometry-based completion methods depend highly on   geometric cues, such as the continuity of local surfaces or volumetric smoothness, which have been applied to retouch small holes on   incomplete point clouds successfully \cite{kazhdan2006poisson, tagliasacchi2011vase, wu2015deep}.
However, it is not applicable for completing missing points of larger regions. Thus, approaches using hand-designed heuristics are proposed to reconstruct surfaces of the 3D objects with a large percentage of missing areas. 
For example, \cite{schnabel2009completion} presented a method to complete 3D shapes with merely partial inputs by combining a series of planes and cylinders. 
Furthermore,  in \cite{li2011globfit}, relations among geometric shapes such as planes and cylinders were proposed to be learned, which is beneficial for improving the performance. For objects with arterial surfaces, in \cite{li2010analysis}, a novel deformable model named arterial snake was proposed, and it successfully captured the topology and geometry simultaneously from arterial objects with noise and large parts missing.

Additionally, in \cite{thrun2005shape, zheng2010non, pauly2008discovering, tevs2014relating, harary2014context},   the symmetry of human-made objects, which usually have structural regularity, was considered. in \cite{thrun2005shape}, the authors identified the probable symmetries and applied them to extend the partial 3D model to the occluded space. \cite{pauly2008discovering} leveraged regular structures that form a lattice with discrete rotational, translational, and scaling symmetries to fill missing regions. \cite{zheng2010non} automatically consolidated and densified real-scan data by detecting repeating structures in input 3D models. \cite{tevs2014relating} sought to quantify the relationship between shapes based on the regularities of symmetric parts. The shape of objects was firstly decomposed into a set of regions, and   a graph was then applied to represent the relations between the regions in terms of symmetric transformations. in \cite{harary2014context}, the authors utilized context information to synthesize geometry that is similar to the remainder of the input objects.
However, all these methods are limited to only completing input point clouds with moderate degrees of missing regions.

\subsection{Template-based Shape Completion}

In addition to the geometry-based methods, some methods follow an alternative strategy, in which they will complete   3D surfaces by deforming or reconstructing   point clouds according to the retrieved, most similar templates from a prepared 3D shape database. These are called template-based methods, which are also known as retrieval-based methods.
As a precondition for the retrieval, a 3D shape database was created in \cite{pauly2005example} to extract   geometric clues for completing missing regions. However, this method  embedding a database retrieval process is time-consuming and labor-intensive since manual interaction is needed to constraint the categories of 3D objects. 
Similarly, the authors in \cite{rock2015completing} proposed a novel completing method automatically for any category of objects based on the use of additional depth images 
  as auxiliary data. An adequate auxiliary database with sufficient elements plays an essential role in the performance of this method.

To avoid the high dependency of large-scale 3D shape databases, some works \cite{schnabel2009completion, nan2010smartboxes, chauve2010robust, li2011globfit, shen2012structure, sung2015data} were proposed to apply geometric primitives in place of a shape database. For example, the authors in \cite{schnabel2009completion} reconstructed missing parts with the guidance of a set of detected primitive shapes (e.g. planes and cylinders). \cite{nan2010smartboxes} presented a novel interactive tool called SmartBoxes to reconstruct  structures that are partially missing from inputs. This allowed the user interactively to fit polyhedral primitives, avoiding an exhaustive search. The authors in \cite{chauve2010robust} plausibly completed   missing scene parts by decomposing 3D space based on planar primitives. The authors in \cite{li2011globfit} sought to simultaneously recover the local missing parts using structural relations from man-made objects, which must include basic primitives. \cite{shen2012structure} presented an assembly approach using predefined geometric primitives to recover 3D structures with a small-scale shape dataset. 
\cite{sung2015data} employed a global optimization method to reconstruct entire surfaces using inference from given geometric information from partial inputs. 

However, such methods exhibit several limitations. First of all, they are not suitable for online applications since the optimization schemes are time-consuming. Secondly, preparing a 3D shape database is labor-intensive since every shape is labeled and segmented manually. At last, they are not always robust to noise or disturbances (e.g., dynamic changes).

\subsection{Learning-based Shape Completion}

Recently, learning-based methods for 3D shape completion have obtained significant developments with the emergence of large-scale 3D synthetic CAD model datasets. 
These state-of-the-art methods have shown excellent performance on various representing formats of 3D models, including voxel grids, point clouds, and meshes. 
Earlier studies of these methods \cite{dai2017shape, song2017semantic, stutz2018learning} involve voxel grids as the representation of 3D shapes since 3D convolution and distance field formats are well suited for processing this kind of discrete and rasterized data. 
\cite{dai2017shape, song2017semantic} are typical examples of voxel-based methods, which adopted a 3D convolutional network to achieve the excellent performance of completing shapes. in \cite{stutz2018learning}, the authors proposed a weakly supervised learning-based method to complete a 3D shape, and this method is easier to achieve 
 in practice.
However, the voxelization representation shows a series of issues. For instance, grid occupancy is predicted independently, which causes   the shape  results to often miss thin structures or contain flying voxels. Moreover, the volumetric representation obscures natural invariance when it comes to geometric transformations and manipulations. In addition, it is computationally expensive to predict volumes of high spatial resolution. 

Thus, some recent works \cite{groueix2018papier, wang2018pixel2mesh, litany2018deformable}, which focus on the completion using a mesh-based representation, have emerged. \cite{groueix2018papier} proposed a novel shape generation network called AtlasNet, which represents a 3D shape as a collection of parametric surface elements. \cite{wang2018pixel2mesh} introduced a graph-based network named Pix2Mesh to reconstruct 3D manifold shapes. \cite{litany2018deformable} explored a variational autoencoder using graph convolutional operations to deformable meshes, which focus on certain objects that undergo non-rigid deformations such as faces or human bodies
. However, they reconstruct   shape information by deforming a reference mesh to a target mesh. 
Therefore, they are not flexible with all typologies. 

In comparison to 3D meshes, or voxels representations, point clouds are a simple structure for the network training procedure. 
In addition, newly created points can easily be added or interpolated to a point cloud since all the points are independent and we do not need to update the connectivity information. 
Some recent work also processes discrete points without structuring via voxels or meshes. 
For example, PCN \cite{yuan2018pcn} was the first approach that directly operated on raw point clouds and outputted complete and dense point clouds robustly with partial inputs. 
Furthermore, TopNet \cite{tchapmi2019topnet} was a tree-structured decoder for point cloud generation. However, they are unable to simultaneously produce evenly distributed and complete point clouds with fine-grained details. In this work, our method falls into this category and builds upon the recent network PCN. Different from PCN, our model can generate more uniform point clouds with fine-grained details.

\section{Overview of methodology}\label{sec: overview}

The point cloud completion task can be regarded as a set problem: given the partial and low resolution points $ {X = {\left \{ P_{i}: i = 1,..., N \right \}}} $, the proposed network VPC-Net aims to generate the complete 3D point cloud $ {Y = F(x) = F(P_{i}: i = 1, ..., N)} $, with \textit{F} being the prediction function. Notably, \textit{X} is not necessarily a subset of \textit{Y} since they can be obtained from a vehicle surface independently. 
\begin{figure*}[!tb]
	\begin{center}
		\includegraphics[width=16cm]{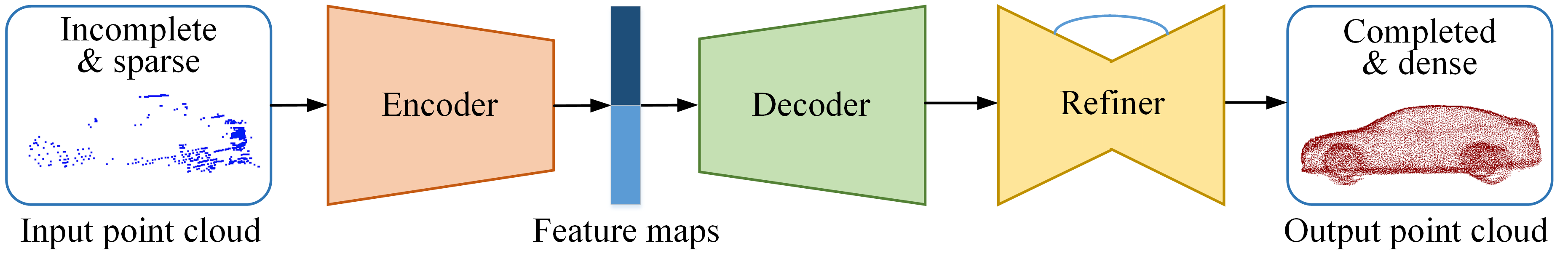}
		\caption{Workflow of the proposed network VPC-Net.}
		\label{fig:Workflow}
    \end{center}
\end{figure*}
The critical architecture of VPC-Net is shown in Fig. \ref{fig:Workflow}, which consists of an encoder module, a decoder module, and a refiner module. Firstly, the encoder is to extract the global features from   raw and sparse point clouds. Secondly, the decoder consists of two parts: (i) it takes the generated global features as input to produce the coarse but complete point cloud, and (ii) it combines the coarse point cloud and global features to generate dense point clouds. Finally, we use the skip connections to concatenate the partial inputs with the previous dense point cloud to preserve the original details. The  refiner further refines the fused 3D point clouds to produce the final completion result. The point clouds generated by VPC-Net should perform three outstanding functions: (i) complete the missing surface with fine-grained structures; (ii) preserve the original details of the inputs; (iii) produce uniform point clouds.

\section{Network architecture}\label{sec: network}

\begin{figure*}[!tb]
	\begin{center}
		\includegraphics[width=16cm]{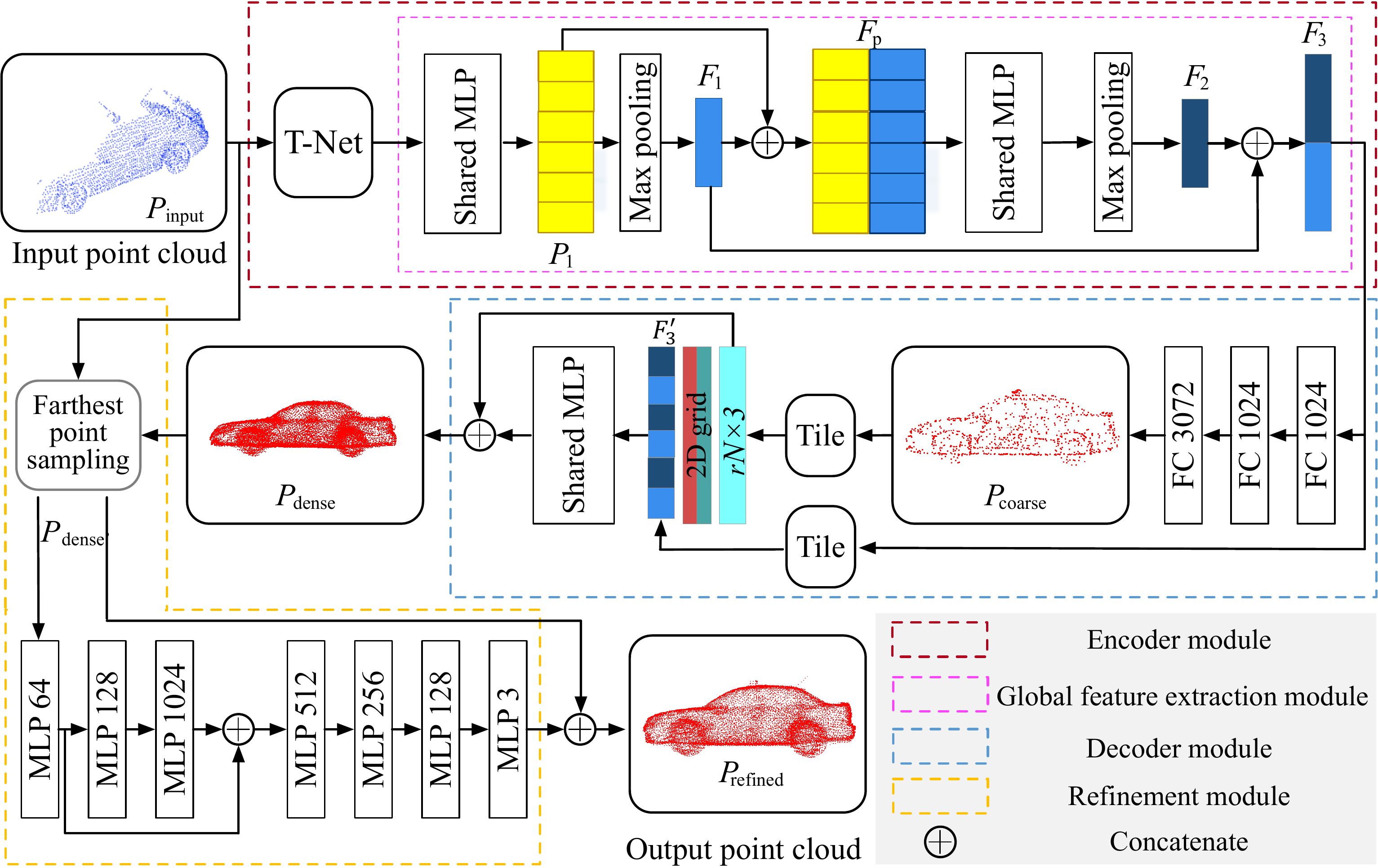}
		\caption{The network architecture of VPC-Net.}
		\label{fig:network}
    \end{center}
\end{figure*}

The architecture of VPC-Net is shown in Fig.~\ref{fig:network}. It includes three sub-networks: feature extraction (encoder), coarse-to-dense reconstruction (decoder), and the refiner. A detailed explanation of each core step will be introduced in the following sections.

\subsection{Encoder}\label{sec:Encoder}

The encoder builds a set of features $ \textit{F} $ for the decoder to estimate the missing surface of 3D vehicles. Therefore, the feature extraction ability of  the encoder plays a vital role in the whole network. If the encoder can effectively combine the local features and global features from the partial inputs, it is significantly beneficial for the 3D coordinates regression of the dense point cloud generation. Our encoder consists of two modules: a spatial transform network module and a global feature extraction module. 
It can be modeled by the combination of two functions,  defined as follows:
\begin{equation} 
    F = Q(P_{input} | w_{Q}), Q = Q_{1}\circ Q_{2}
\label{Eq:encoder}
\end{equation}
\noindent 
where $Q_1$ and $Q_2$ are the spatial transform network and the global feature extraction module, respectively.   $w_{Q}$ denotes the weight parameters of $ Q $, and $ P_{input} $ is the partial input point cloud.

\subsubsection{Spatial transform network}\label{subsec:STN}

Since the input point clouds of the vehicle are disordered and their poses are diverse, this will cause difficulty in accessing the unified features for neural networks. Therefore, in order to facilitate the extraction of features, we hope the input point clouds have a neat pose. In other words, learned features from input point sets should be invariant to geometric transformations. 

Aligning all point sets to a canonical space is a natural solution. In \cite{jaderberg2015spatial},  the authors used a spatial transformer for learning invariance to translation and rotation in 2D images. Inspired by this, we adopt a 3D spatial transform network named T-Net \cite{qi2017pointnet} to predict a $\textit{3} \times \textit{3}$ transformation matrix for the original point clouds. Furthermore, we directly multiply this transformation matrix and the coordinates of input points. Thus, the inputs are aligned to a canonical space so that the following network can attentively learn a unified and standardized feature. 

T-Net is like a mini-PointNet \cite{qi2017pointnet}, which includes a shared Multiple Layer Perception (MLP) network, a max-pooling layer, and two fully connected layers. It takes the raw point clouds as inputs and outputs a $\textit{3} \times \textit{3}$ matrix. In detail, the MLP network first encodes each point to multiple dimensions $ {[64, 128, 1024]} $.   A max-pooling layer is adopted and followed by two fully connected layers with output sizes $ {[512, 256]} $. The regressed matrix is initialized as an identity matrix. Except for the last layer, all layers are followed by a ReLU activation and a batch normalization layer. 

\subsubsection{Global feature extraction module}\label{subsec:Point feature extraction module}

Generally, our global feature extraction module is based on recently advanced feature extraction network PointNet \cite{qi2017pointnet}, which directly operates on point clouds. Inspired by this, the encoder, as illustrated in Fig.~\ref{fig:network}, adopts two stacked PointNet layers to extract the geometric information for the input point cloud. 
Each PointNet layer comprises   one shared MLP and one max-pooling layer as a basic module. 
In the first PointNet layer, we learn a point-wise feature $ {P_{1}} $ from the points of ${N_{input}} \times \textit{3}$ transformed by the STN, where $ {N_{input}}$ is the amount of points and \textit{3} is the \textit{x,y,z} coordinates of each point. 
Afterwards, a max-pooling layer is employed on $ {P_{1}} $ to output a 256-dimensional local feature vector $ {F_{1}}$. 
In the second PointNet layer, we firstly concatenate the local latent space with every independent point feature by feeding $ {F_{1}}$ back to the point-wise feature $ {P_{1}} $. 
The  global latent vector ${F_{2}}$ is then extracted from the aggregated point features ${F_{p}}$ through the second PointNet layer, with the size ${F_{2}:=1024}$. 

However, it always loses the fine details of the inputs since the latent space extracted by the last max-pooling layer only represents the rough global shape. Inspired by the skip connection from U-Net \cite{ronneberger2015u}, we design a point feature enhancement (PFE) layer, which concatenates the global feature ${ F_{2}} $ with the local feature $ {F_{1}} $ to synthesize the final feature space $ {F_{3}} $. Size $ {F_{3}:=1280} $, and it includes both low-level and high-level feature information. Experimental results in Section \ref{sec:Ablation study} show that this design improves the feature extraction ability of  the encoder for partial inputs.

\subsection{Decoder}\label{sec:decoder}

The decoder is responsible for converting the final global latent vector $ {F_{3}} $ into dense, evenly, and complete 3D point clouds. In this stage, a coarse-to-fine completion strategy is applied for generating the 3D coordinates of point clouds. Inspired by 3D object reconstruction network RealPoint3D \cite{xia2019realpoint3d}, we explore three fully connected layers to generate a sparse point cloud with a complete geometric surface. Lastly, it outputs the final vector with $ {3N} $ units, and we reshape it into an $ {N}\times{3}$ coarse point cloud $ {P_{coarse}} $.
\begin{figure}[ht]
	\begin{center}
		\includegraphics[width=8cm]{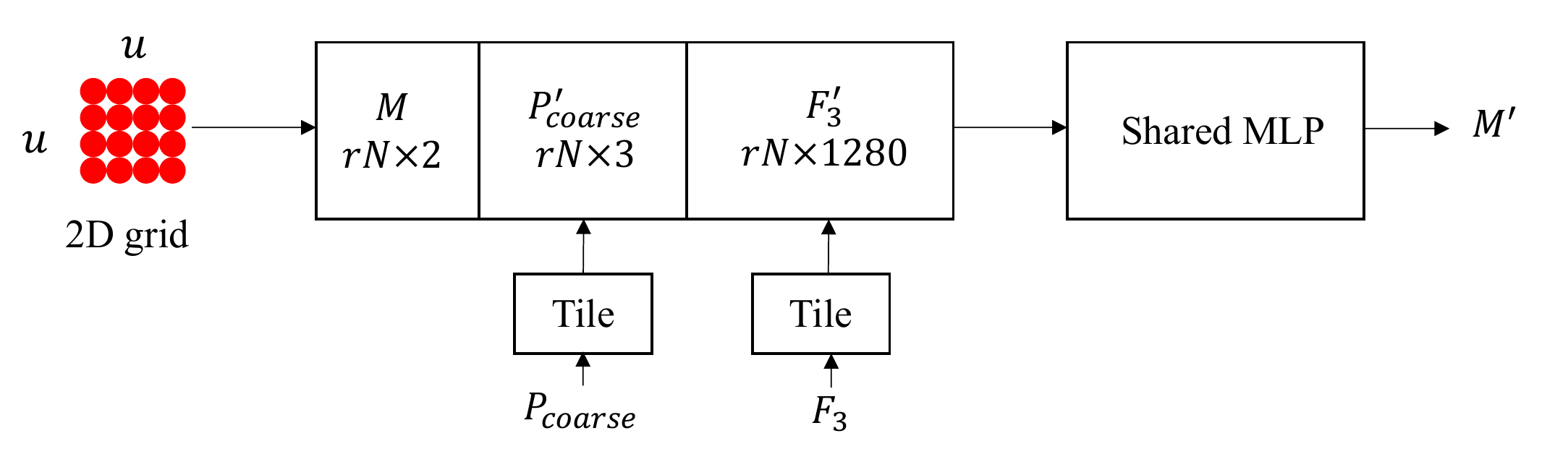}
		\caption{The detailed concatenating operation in the decoder.}
		\label{fig:decoder}
    \end{center}
\end{figure}
\vspace{-0.2cm}
However, the fully connected layer is not suitable for generating dense points. It will cause some points to be over-concentrated when we regress a large number of points. The reason is that fully connected layers are not restrictive on the local density. Therefore, in the second stage, we first tile the points in $ {P_{coarse}} $ to produce a dense point set $ P_{coarse}^{'}:= {rN}\times{3} $, where $ {r} $ is the up-sampling rate. We then apply a folding-based operation \cite{yang2018foldingnet} to deform a unique 2D grid vector and concatenate it with each point of the coarse point cloud to obtain new patches. This operation can increase the difference between the duplicated points. In other words, we regard each point of the coarse point cloud as spatial keypoints and take them as center points to generate a series of surrounding points. To make full use of the features of input point clouds, we concatenate the points in $ P_{coarse}^{'} $,  the tiled global feature space $ F_{3}^{'} $, and the deformed 2D grids to obtain a new aggregated feature. The detailed concatenating operation is shown in Fig.~\ref{fig:decoder}. The coordinates of points on a zero-centered $ u \times u $ grid ($ u^{2} = {rN}\times{3} $) are first deformed into a ${rN}\times{2} $ matrix $ M $~\cite{yang2018foldingnet}. We then concatenate $ M $ with the coordinates of the tiled coarse point cloud $ P_{coarse}^{'} $ and the duplicated global feature vector $ F_{3}^{'} $. Furthermore, the aggregated feature is passed through a shared MLP with sizes $ {[512, 512, 3]} $ to generate a new ${rN}\times{3}$ matrix $ M^{'} $. This shared MLP can be regarded 
 as a non-linear mapping that transforms the 2D grid into a smooth 2D manifold in 3D space \cite{yuan2018pcn}. Finally, the dense point cloud $ P_{dense}:= {rN}\times{3} $ is generated by adding the coordinates of each point in $ P_{coarse}^{'} $ to the matrix $ M^{'} $. 
 

\subsection{Refiner}\label{sec: refiner}

Although the decoder can produce impressive results, it always loses the fine-grained details of the inputs, and the points are unevenly distributed. To tackle these problems, we combine the partial inputs $ P_{input} $ with the outputs $ {P_{dense}} $ generated by the decoder. This operation can fully retain the details of the input point cloud. However, the linear combination will cause the merged points to be non-uniform since the two point clouds have a different density and there may be overlap  between them. Thus, we apply   farthest point sampling (FPS) to sample a uniform distributed subset point cloud $ P_{dense}^{'} $  with a size of $ {rN}\times{3} $. The hyperparameter $r$ is the same in the decoder and the refiner. In this work, $ r = 16 $.

The refiner can be regarded as a point feature residual network. We hope the refiner can predict per-wise offsets $ {o_{x}}, {o_{y}}, {o_{z}}$ for every point in $ P_{dense}^{'} $. Therefore, we pass the points $ P_{dense}^{'} $ through a series of MLPs to predict point feature residuals since neural networks are better at residuals \cite{wang2018pixel2mesh}. Specifically, inspired by the structure of  an encoder-decoder network, we adopt a bottom-up and top-down strategy to refine  the point coordinates. The refiner consists of seven MLPs. It first encodes each point into multiple dimensions $ {[64, 128, 1024]} $. We then decode it to generate the offsets of each point with dimensions of $ {[512, 256, 128, 3]} $. 
Expect for the last layer, followed by a batch normalization layer and a tanh activation, other MLPs are followed by a batch normalization layer.
In addition, we hope that the local feature can be preserved in the following layers. Thus, we combine the feature with dimensions of $ {64} $ and the bottleneck layer with a size of $ {1024} $, as shown in Fig.~\ref{fig:network}.
Overall, in this refiner, the final generated point sets $ P_{dense}^{'} $ is defined as
\begin{equation}
    P_{refined} = R(P_{dense}^{'}) + P_{dense}^{'}
\end{equation}
\noindent
where $ R\left \{ . \right \} $ predicts per-wise displacements by the refiner.

\subsection{Loss function}\label{sec: loss}

The loss function of our network is defined as the topological distance between the completed object and the ground truth. Inspired by \cite{fan2017point}, we adopt the Chamfer Distance (CD) and Earth Mover's Distance (EMD) to optimize the network. Distance metric functions are highly efficient and invariant to permutations of the relative ordering of points.
The CD between the completed point cloud $ P_c $ and the ground truth $ P_{gt} $ is defined as
\begin{equation} 
\begin{split}
    d_{chamfer}(P_c,P_{gt})=\sum_{x\in P_c}\min \limits_{y\in P_{gt}}{\parallel x-y \parallel}_2^2 \\
    + \sum_{x\in P_{gt}} \min \limits_{y\in P_c}{\parallel x-y \parallel}_2^2
\end{split}
    \label{Eq:C_distance}
\end{equation}
\noindent 
where $ P_c,P_{gt}\subseteq R^3 $. Intuitively, it aims to find the closet neighbor between the two point sets in two directions. Each point of $ P_c $ is mapped to the closet point in $ P_{gt} $, and vice versa. Thus, the size of $ P_c $ and  $ P_{gt} $ is not required to be the same. It is a computationally light function with $ O(nlogn) $ complexity for the nearest neighbor search. However, it is a problematic metric since it cannot ensure the uniformity of predicted points \cite{mandikal2019dense}. In addition, it is sensitive to the detailed geometry of outliers \cite{tatarchenko2019single}. To alleviate these problems, the EMD between $ P_c $ and $ P_{gt} $ is proposed by
\begin{equation} 
    d_{EMD}(P_c,P_{gt})=\min \limits_{\phi :P_c \to  P_{gt} }\sum _{p\in P_c}{\parallel p-\phi (p) \parallel}_2
\label{Eq:E_distance}
\end{equation}
\noindent 
where $ P_c,P_{gt}\subseteq R^3 $, $\phi :P_c \to  P_{gt} $ is a bijection. Unlike CD,   the size of $ P_c $ and $ P_{gt} $ must be the same since it is a point-to-point mapping function. However, it has a major drawback:   the $ O(n^{2}) $ computing complexity is too expensive. It is not suitable for predicting dense points in the network. 

Therefore, we propose a training strategy that can take advantage of both the distance functions. To make sure the generated coarse point cloud is even  and has general geometry, we apply the EMD loss for $ P_c $ predicted by the encoder. The predicted dense point clouds $ P_{dense} $ and $ P_{dense}^{'} $ are optimized via the CD loss. More formally, the overall loss is defined as 
\begin{equation}
    \begin{split}
    L(P_{coarse}, P_{dense}, P_{dense}^{'}, P_{gt})
    &= d_{EMD}(P_{coarse},\tilde{P_{gt}}) \\
    & + \gamma d_{chamfer}(P_{dense},P_{gt}) \\
    & + \beta d_{chamfer}(P_{dense}^{'},P_{gt}) 
    \end{split}
\end{equation}
\noindent where $\tilde{P_{gt}}$ is the subsampled ground truth with the same size as $P_{coarse}$.  $\gamma$ and $ \beta $ are hyperparameters to balance their relationship.

\section{Experiments} \label{sec:experiments}

In this section, we performed experiments to demonstrate the effectiveness of the proposed VPC-Net when completing point clouds of real LiDAR scans. We will first introduce the experimental datasets and the generation of training data in Section \ref{sec:experimental datasets}. In Section \ref{sec: evaluation metrics},  we will describe the evaluation metrics used for assessing the performance of VPC-Net, as well as baseline methods. Furthermore, the implementation details and training processing will be introduced in Section \ref{sec:implementation}. Finally, experimental results are presented in Section~\ref{sec:Experimental results}.

\subsection{Experimental datasets}\label{sec:experimental datasets}

In the experiments, we tested our proposed VPC-Net method on three different datasets, including  the ShapeNet dataset \cite{chang2015shapenet},  the KITTI dataset \cite{geiger2013vision}, and  the TUM-MLS-2016 dataset \cite{zhu2020tum}.

\subsubsection{ShapeNet dastaset} 
\begin{figure*}[ht]
	\begin{center}
		\includegraphics[width=1.0\textwidth]{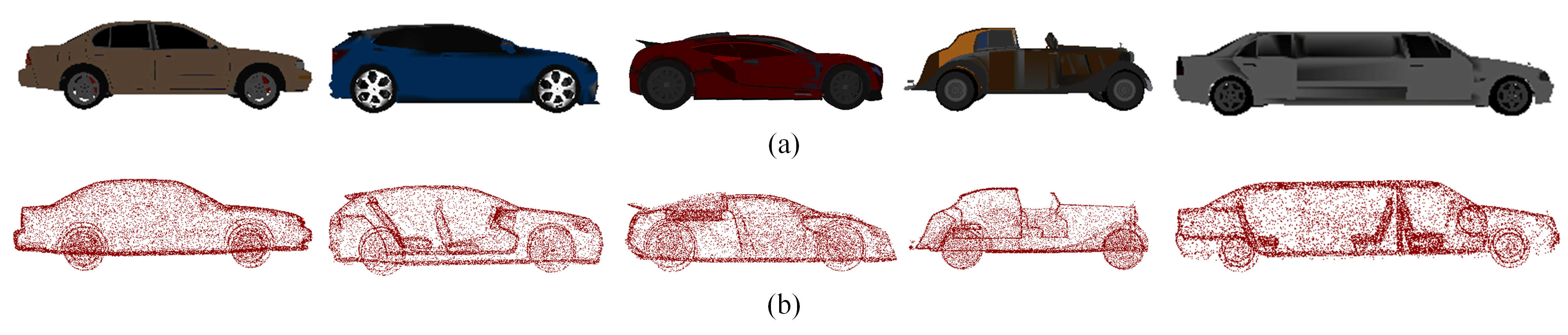}
		\caption{Examples of CAD models and sampled point clouds of vehicle instances from the ShapeNet dataset. (a) CAD models of vehicle instances stored in ShapeNet. (b) Generated complete point clouds sampled uniformly from these CAD models.}
		\label{fig:shapenet_examples}
    \end{center}
\end{figure*}
\begin{figure*}[ht!]
	\begin{center}
		\includegraphics[width=1.0\textwidth]{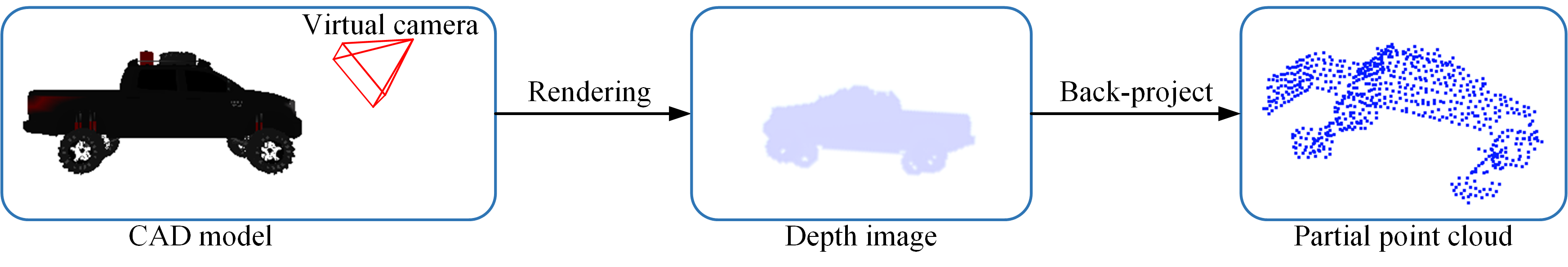}
		\caption{The pipeline of partial input generation.}
		\label{fig:generation_part}
    \end{center}
\end{figure*}
ShapeNet \cite{chang2015shapenet} is a richly annotated and large-scale 3D synthetic dataset, which covers 220,000 CAD models and 3135 categories of objects. 
In this work, we use synthetic CAD models on the category of cars from ShapeNet to create a vehicle dataset containing pairs of partial and complete point clouds, in order to train our model. 
Specifically, it includes a total of 5677 different instances of vehicles, which are split into training data, validation data, and test data. 
Among them, 100 instances are used for validation, and 150 instances are utilized for testing. The remaining instances are reserved for training. For creating complete point clouds as ground truth, for each CAD model of a vehicle instance, 16,384 points are sampled uniformly on the surface of each CAD model of a vehicle as the synthetic point cloud. 
Fig.~\ref{fig:shapenet_examples} shows examples of complete point clouds of vehicle instances from CAD models in ShapeNet. 
Instead of using subsets of complete point clouds as partial inputs, we rendered the CAD models of vehicle instances to a set of depth images from a variety of view angles and then back-projected these depth images to different view planes to generate partial point clouds. This operation can make the incompleteness distribution of partial point clouds closer to real-scan data.

Following the data generation in PCN~\cite{yuan2018pcn}, we illustrate the pipeline of generating partial inputs from the ShapeNet dataset in Fig.~\ref{fig:generation_part}.
The depth images are generated by placing a virtual RGB-D camera at different view angles. The camera is designed to be oriented towards the center of the 3D model. 
We then randomly select a series of viewpoints only to generate incomplete shape 
  scans 
 obtained through limited view access. Lastly, the resulting depth maps are back-projected to form partial point clouds.
In this work, we chose eight randomly distributed viewpoints to generate eight partial point clouds for each training 3D CAD model of a vehicle. 
Notably, the resolution of these partial scans can be different.
The reason for generating training point clouds from a synthetic 3D dataset is that it consists of a wide variety of complete and detailed 3D vehicle models, while they are not available in real-scanned LiDAR datasets. 
Moreover, scanning thousands of vehicles using LiDAR systems for acquiring complete point clouds as the ground truth is quite time-consuming and labor-intensive, which is not a practical solution.
Recently, some high-quality 3D reconstruction datasets have emerged such as ScanNet \cite{dai2017scannet} and S3DIS \cite{armeni2017joint}, which can also provide training data with high quality. However, they are mainly focused on indoor scenes, not including any objects in outdoor scenarios.

\subsubsection{KITTI dataset}
\begin{figure*}[!tb]
	\begin{center}
		\includegraphics[width=16cm]{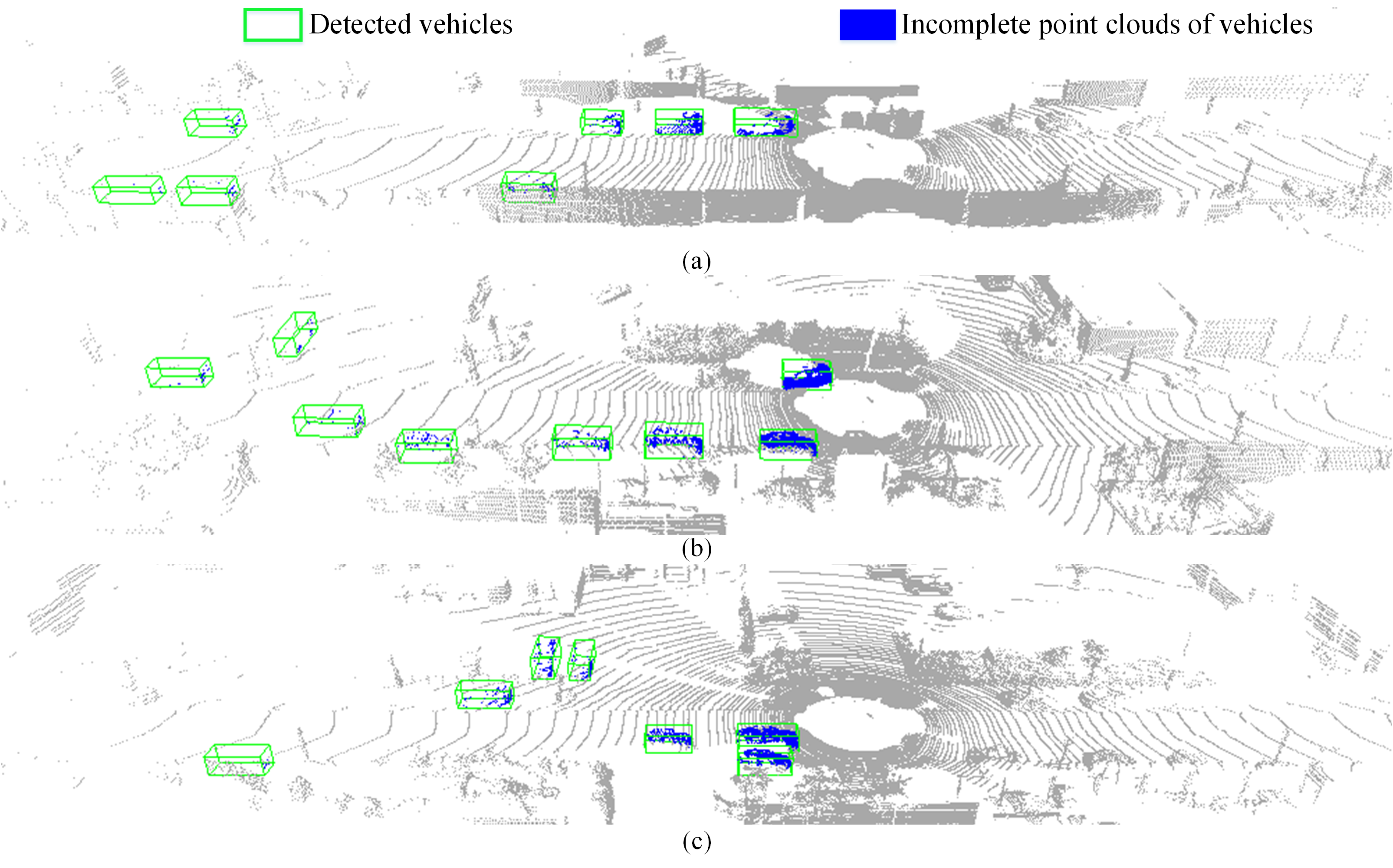}
		\caption{Example frames of  the  'City' category from the KITTI dataset. The vehicle points, background points, and bounding boxes are shown in blue, gray, and green colors, respectively.}
		\label{fig:kitti_dataset}
    \end{center}
\end{figure*}
The KITTI \cite{geiger2013vision} dataset provides raw point clouds collected by  the Velodyne HDL-64E rotating 3D laser scanner and annotations for   vehicle instances in the form of 3D bounding boxes. 
It records six hours of traffic scenarios, which are diverse and capturing real-world traffic situations with many static and dynamic vehicles. 
The raw dataset includes five categories of objects, namely 'Road', 'City', 'Residential', 'Campus', and 'Person'. 
In the data category 'City', it is composed of about 28 sequences (i.e., 8477 frames). In each sequence of the raw data, apart from objects annotated with 3D bounding boxes, tracklets and calibration are also provided. 
Three example frames in 'City' are shown in Fig.~\ref{fig:kitti_dataset}.
As seen in Fig.~\ref{fig:kitti_dataset},  the major challenge of this dataset is twofold. 
One is that the point clouds of vehicles are very sparse and exhibit a significant loss of content, while another is that target vehicles appear in an arbitrary location with variable sizes.
In this work, we took one sequence in the category 'City' from the KITTI dataset as experimental data. Specifically, we extracted 2483 partial point clouds of vehicles from every frame based on their bounding boxes. 

\subsubsection{TUM-MLS-2016 dataset}

\begin{figure*}[!tb]
	\begin{center}
		\includegraphics[width=1.0\textwidth]{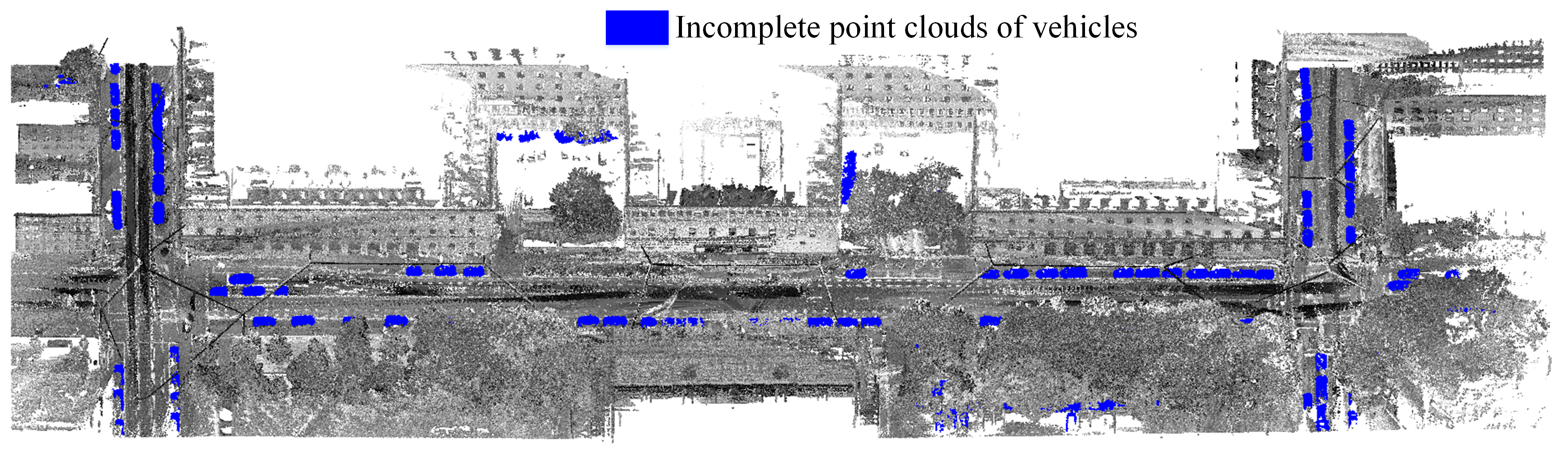}
		\caption{Point clouds of Arcisstrasse from the TUM-MLS-2016 dataset. The vehicle points and background points are shown in blue and gray colors, respectively.}
		\label{fig:tum_dataset}
    \end{center}
\end{figure*}

TUM-MLS-2016 \cite{zhu2020tum} is a mobile laser scanning dataset covering around 80,000 $m^2$ with annotations. This dataset was acquired by Fraunhofer Institute of Optronics, System Technologies and Image Exploitation (IOSB), via two Velodyne HDL-64E laser scanners and then annotated by the Chair of Photogrammetry and Remote Sensing of TUM. 
Unlike the KITTI dataset, the TUM-MLS-2016 dataset provides an aggregated point cloud of the whole obtained sequence. 
It covers an urban area with approximately 1 $km$ long roadways and includes more than 40 million annotated points with labels for 
eight classes of objects.
In Fig.~\ref{fig:tum_dataset}, we give an illustration of scanned vehicles on the Arcisstrasse of this dataset.
We extracted point clouds of vehicles as the testing data based on the provided annotations of parked vehicles. 
As shown in Fig~\ref{fig:tum_dataset}, the point clouds of vehicles in the TUM-MLS-2016 dataset are denser than in those in the KITTI dataset. They are also incomplete, although the missing content  is  less severe.

\subsection{Evaluation metrics} \label{sec: evaluation metrics}

The performance of our proposed method is evaluated by two commonly applied metrics: CD (see Eq.~\ref{Eq:C_distance}) and EMD (see Eq.~\ref{Eq:E_distance}), between the completed point cloud and  the ground truth. 
The definitions of CD and EMD have been given in Section~\ref{sec: loss}. 
For computing the metrics with a lower computational cost~\cite{kang2020deep}, we normalized the dimensions of both the ground truth and completed point clouds, by regarding the length of the bounding box of length as one unit.

\subsection{Implementation details and training process}\label{sec:implementation}
\begin{figure*}[!tb]
	\begin{center}
		\includegraphics[width=16cm]{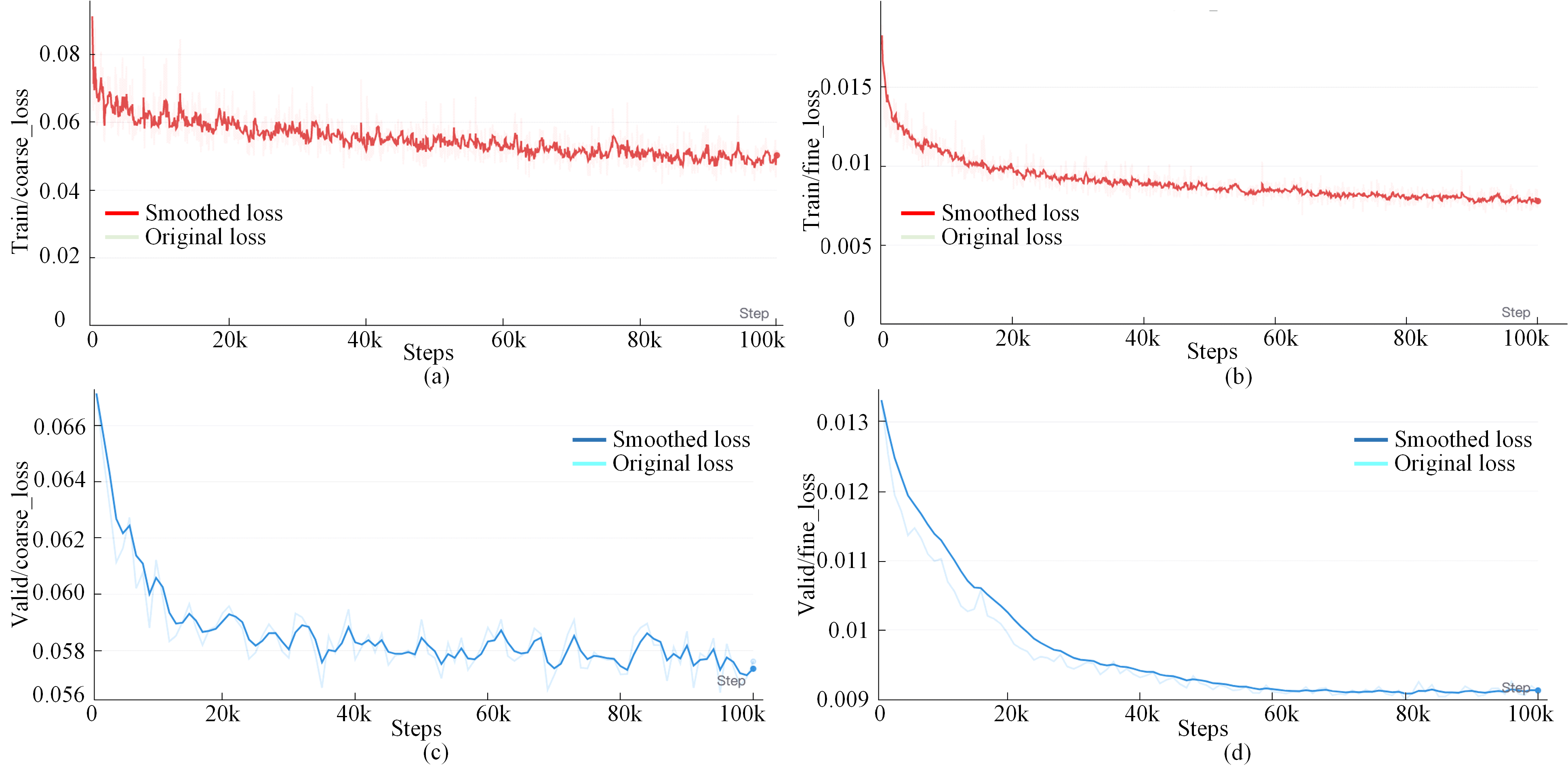}
		\caption{Visualization of the training process. EMD errors for coarse point cloud generated by the decoder in (a) the training stage and (b) the validation stage. CD errors for dense point cloud produced by the refiner in (c)  the training stage and (d)  the validation stage.}
		\label{fig:learning_curve}
    \end{center}
\end{figure*}
\vspace{-0.25cm}
The proposed network VPC-Net was implemented in the Tensorflow framework and trained on a single NVIDIA Titan Xp GPU with 12 GB of memory. 
In the training stage, we set the batch size to eight. The Adam optimizer was used in the models for 100 K steps. 
The size of the coarse output generated by the encoder was 1024. 
The initial learning rate was set to 0.0001. 
They 
 were decayed by 0.7 after every 50 K steps and clipped by $10^{-6}$. $\gamma$ and $ \beta $ were  made equal. 
They gradually increased from 0.01 to 1 in the first 50 K steps.
Notably, the resolutions of the inputs were various, from a few hundred points to thousands of points. 

Additionally, to demonstrate the training process more vividly, we plotted the learning curve of the proposed method VPC-Net (see Fig.~\ref{fig:learning_curve} for illustration). 
The training losses and validation losses both consisted of two different types of losses. 
One is the CD for generated coarse point clouds, while the other is the EMD  for the produced dense point clouds. 
As shown in Figs.~\ref{fig:learning_curve}a and \ref{fig:learning_curve}b,  the training losses gradually decreased as the number of training steps increased and converged until 100 K steps. 
The validation losses are shown in Figs.~\ref{fig:learning_curve}c and \ref{fig:learning_curve}d, which also prove the proposed method VPC-Net converges at 100 K training steps. 
\begin{table*}[ht!]
    \centering
    \caption{Quantitative comparison (smaller value represents better performance) of our method against the state-of-the-art methods on ShapeNet.}    
    \begin{tabular}{|c|c|c| }
        \toprule
        Methods  & Mean Chamfer Distance per point $(10^{-3})$ & Mean Earth Mover's Distance per point $(10^{-2})$   \\
        \midrule
        3D-EPN~\cite{dai2017shape}      & 22.308 & 10.7080 \\
        PCN~\cite{yuan2018pcn}     & 11.668 & 6.0480\\
        TopNet~\cite{tchapmi2019topnet}  & 13.765 & 9.6840 \\
        VPC-Net  & \textbf{8.662} & \textbf{5.1677} \\
        \bottomrule        
    \end{tabular}
    \label{tab:Tab1}
\end{table*}

\subsection{Experimental results}\label{sec:Experimental results}
\begin{figure*}[!tb]
	\begin{center}
		\includegraphics[width=1.0\textwidth]{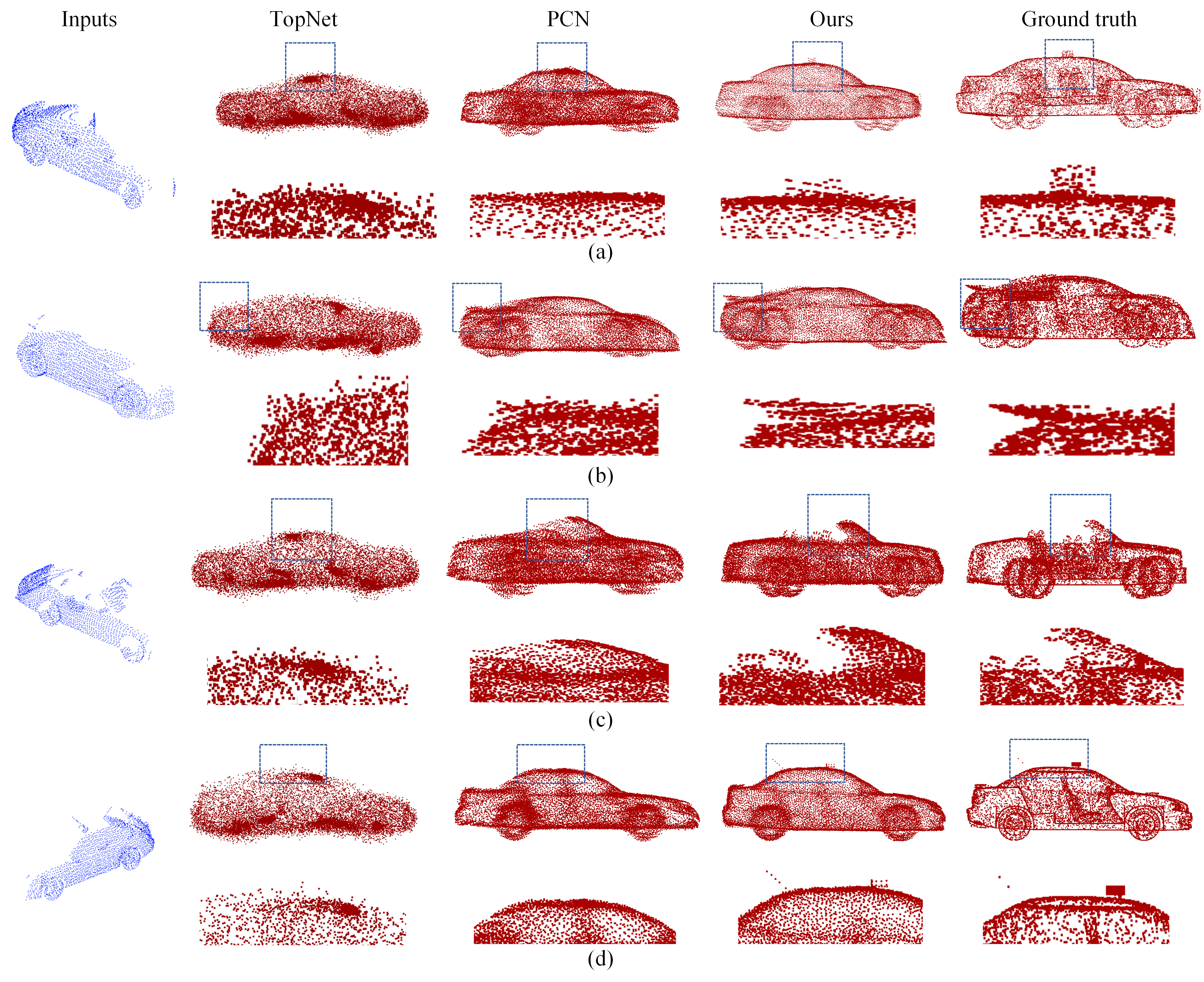}
		\caption{Qualitative comparison of our method against other state-of-the-art methods on ShapeNet. (a)-(d) are four different testing examples. From left to right: input partial point clouds, TopNet \cite{tchapmi2019topnet}, PCN \cite{yuan2018pcn}, VPC-Net, and ground truth.}
		\label{fig:shapenet}
    \end{center}
\end{figure*}
\begin{figure*}[!tb]
	\begin{center}
		\includegraphics[width=16cm]{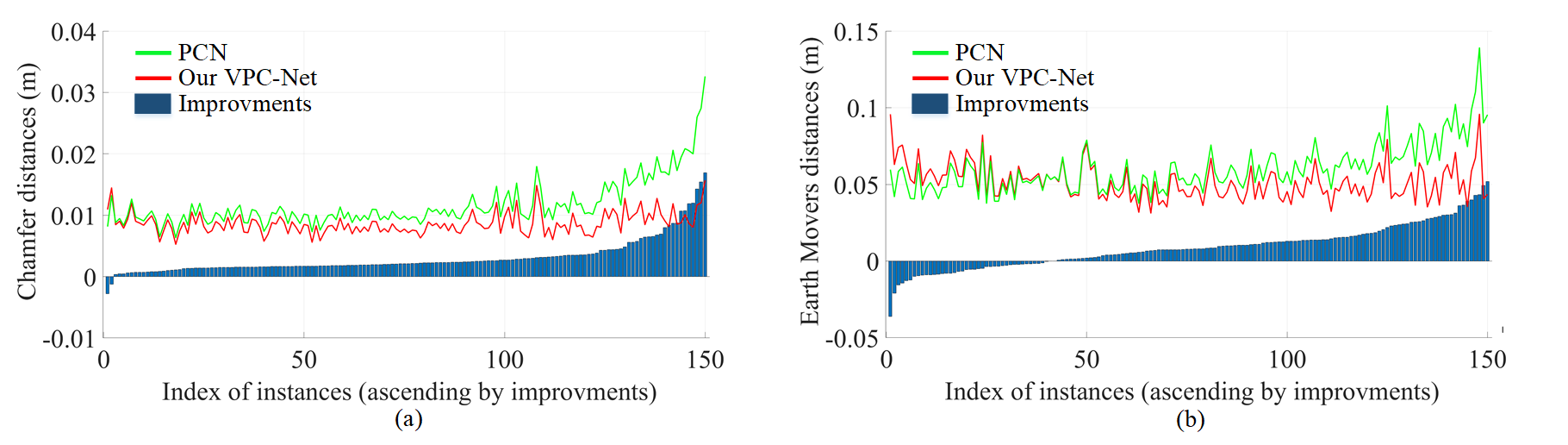}
		\caption{Comparisons between the quantitative results of (a) PCN \cite{yuan2018pcn} and (b) VPC-Net.}
		\label{fig:comparision_pcn}
    \end{center}
\end{figure*}

\begin{figure*}[!tb]
	\begin{center}
		\includegraphics[width=16cm]{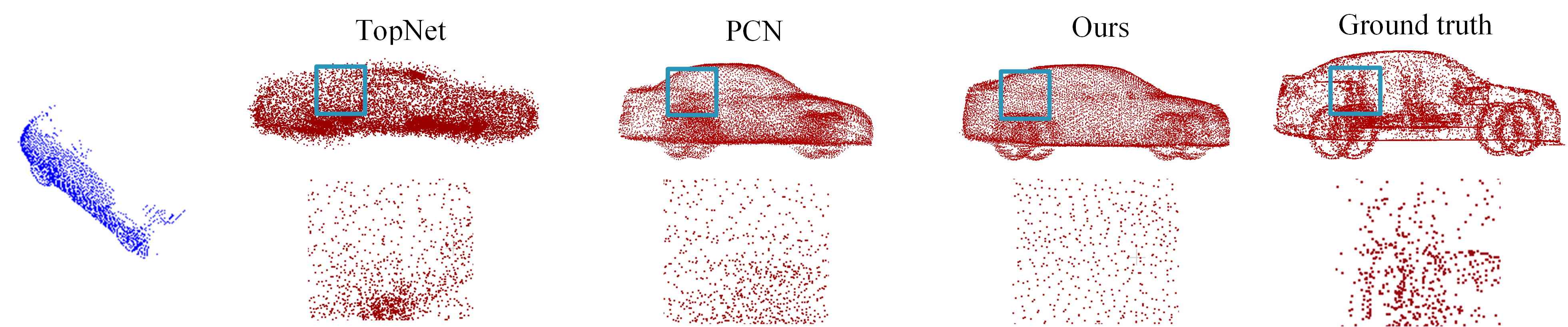}
		\caption{Example point cloud in the same area with different completion methods. From left to right: input partial point clouds, TopNet \cite{tchapmi2019topnet}, PCN \cite{yuan2018pcn}, VPC-Net, and ground truth.}
		\label{fig:uniform}
    \end{center}
\end{figure*}
\subsubsection{Point completion on the ShapeNet dataset}\label{sec: results on ShapeNet}

For evaluating the performance of our proposed method in completing point clouds of synthetic models, we compared our approach against the following state-of-the-art methods on the ShapeNet testing data, including 3D-EPN \cite{dai2017shape}, PCN \cite{yuan2018pcn}, and TopNet \cite{tchapmi2019topnet}.
3D-EPN \cite{dai2017shape} is a typical volumetric completion method, which was trained on the large-scale synthetic dataset as well. 
PCN \cite{yuan2018pcn} is a pioneering method that completes partial inputs using point clouds directly, which conducted   end-to-end training through an auto-encoder. 
TopNet \cite{tchapmi2019topnet} is the newest end-to-end point cloud completion method. 
For a fair comparison, all methods were trained and tested on the same data for all experiments. 
The size of the output point cloud and the ground truth was fixed to 16,384 points. 
Quantitative and qualitative results are shown in Table \ref{tab:Tab1} and Fig.~\ref{fig:shapenet}, respectively.    

Table \ref{tab:Tab1} shows that our proposed VPC-Net outperforms other methods significantly. 
In this table, the value of CD and EMD metrics are scaled by 1000 and 100, respectively.
We obtained a relative improvement on the average CD value by 25.7$\%$ and the average EMD value by 14.6$\%$ over the second-best approach PCN.
Note that the values of EMD are much higher than those of CD. 
The reason is that EMD is a one-to-one distance matching metric, whereas CD can have one-to-many correspondences between   points. 
A visualization of  the point cloud completion results for different methods is shown in Fig.~\ref{fig:shapenet}. 
Comparing the results generated by our method and by others, we can observe that VPC-Net can produce more uniform point clouds with more fine-grained details, while others fail to recover such structures.
Particularly, from the below-up views, it is clearly seen that our method can preserve fine details in the completed results, such as the signboard on the roof of the taxi (see Fig.~\ref{fig:shapenet}a, car spoilers (see Fig.~\ref{fig:shapenet}b), and the antenna of cars (the last row (see Fig.~\ref{fig:shapenet}d)). 

To better display the more specific performance, in Fig.~\ref{fig:comparision_pcn}, we visualize the extent of  the improvement of our results over the second-best approach PCN on CD and EMD for all instances in the test dataset. In this figure, the horizontal axis indicates different vehicles. 
The height of the blue bar represents the increased value of VPC-Net over PCN. The green curve is the error of PCN, and the difference between the green curve and the blue bar is the error of VPC-Net.
It can be seen that our proposed method is a  significant improvement for the majority of shown instances. 
In addition,  our method achieves the greatest improvement in instances that PCN   provides highly noisy results, illustrating that our method is able to handle these challenging examples where previous methods fail. Moreover, to demonstrate that the points completed by VPC-Net are much more uniform than those generated by other baseline methods, three patches of spots in the same area produced by different methods are shown in Fig.~\ref{fig:uniform}. 
From the blown-up views of Fig.~\ref{fig:uniform}, we can see that both TopNet and PCN have heavily cluttered regions, while our completion is more evenly distributed.
We can also observe that TopNet tends to generate several subsets of clustered points. 
PCN cannot ensure the uniformity of local distribution of points. 
The global distribution of points in our outputs is   greater  than those of TopNet and PCN. 

The uniformity and fine-grained details in our completions can be attributed to two factors: 
(1) We adopted two loss functions at two stages of the network. 
At the first stage of generating coarse point clouds, EMD loss forces the predictions to be uniform. 
Thus, the produced dense point clouds tend to be even since the following stage is trained to upsample the coarse predictions, despite using the CD as the loss. 
(2) The refiner first adopts the FPS method to sample the aggregated point cloud by concatenating the partial inputs and the dense outputs from the decoder and then uses residual networks to refine it. 
This operation preserves the details of inputs and guarantees certain degrees of uniformity.

\subsubsection{Point completion on the KITTI dataset}\label{sec: results on kitti}
\begin{figure*}[ht]
	\begin{center}
		\includegraphics[width=16cm]{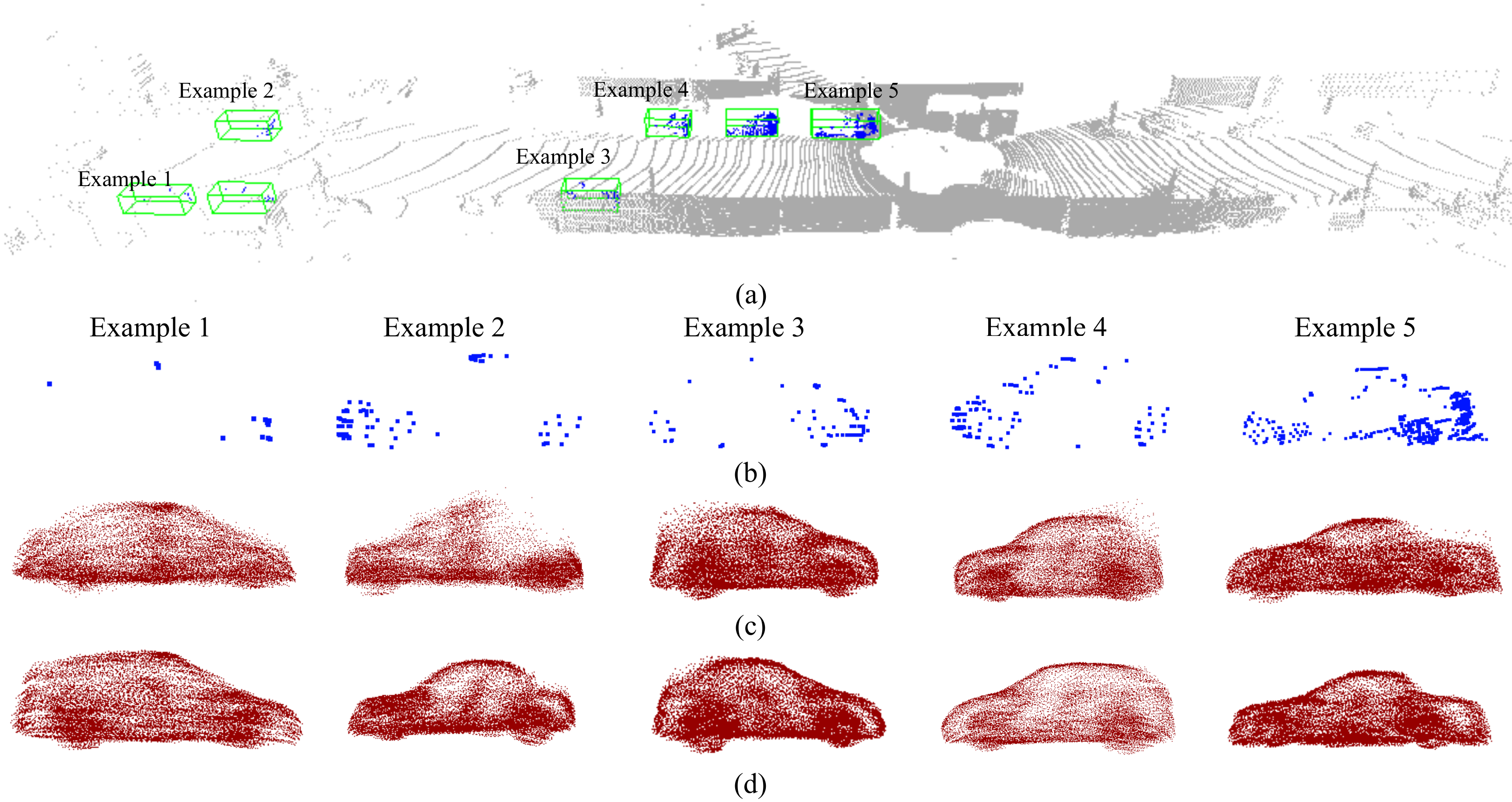}
		\caption{Completed 3D point clouds using real-scan data from the KITTI dataset. (a) Five detected vehicle examples in a single frame. (b) Partial point clouds. (c) Completed point clouds by PCN \cite{yuan2018pcn} and (d) completed point clouds by VPC-Net.}
		\label{fig:kitti}
    \end{center}
\end{figure*}
For evaluating the performance of our method on real scan LiDAR data, we tested our method for point cloud completion on the KITTI dataset.

We extract 2483 partial point clouds of cars from every frame based on their bounding boxes. 
Each extracted point cloud was transformed into the bounding box's coordinate system and then completed by our method trained on the ShapeNet dataset. 
Lastly, we turned them back to the world coordinates. 
Considering the extra noisy points from the ground or nearby objects   within the car’s bounding box, we removed the FPS operation in the refiner since it would bring this noise into the final completed results. 
Note that there are no ground truth point clouds in this dataset.

The qualitative results are shown in Fig.~\ref{fig:kitti}. 
We visualized the single frame raw data and chose five detected vehicles as the testing data, as shown in Fig.~\ref{fig:kitti}a. Fig.~\ref{fig:kitti}b shows five sparse and partial input point clouds, while Figs.~\ref{fig:kitti}c and \ref{fig:kitti}d display the completed point clouds by PCN and our method, respectively. 
From Fig.~\ref{fig:kitti}, we can see that VPC-Net has a better generalization capability and has complete shapes that show that the point sets are evenly distributed on the vehicle surface. Note that both networks were trained on the same ShapeNet training set and tested on KITTI.
For example, for Example 2 in Fig.~\ref{fig:kitti}, the result generated by our method includes the details of missing parts, and all points are more evenly distributed on the geometric surface, while point sets completed by PCN are messy and lose detailed structures of the rear of the car. 
We can also see that many points from PCN escaped the car surface, which can be observed in Example 3, Example 4, and Example 5.

Based on the obtained outputs and comparisons, we can conclude from Fig.~\ref{fig:kitti} that our method is robust to different resolutions of input point clouds, which is an essential characteristic for handing real scan data. 
For example, the point clouds of Examples 1 and 3 have 12 and 100 points, respectively, while   903 points are included in the case of Example 5.
In spite of this, our method is able to produce uniformly, dense, and complete point clouds with finely detailed structures. 

\subsubsection{Point completion on the TUM-MLS-2016 dataset}\label{sec: results on tum dataset}
\begin{figure*}[ht]
	\begin{center}
		\includegraphics[width=16cm]{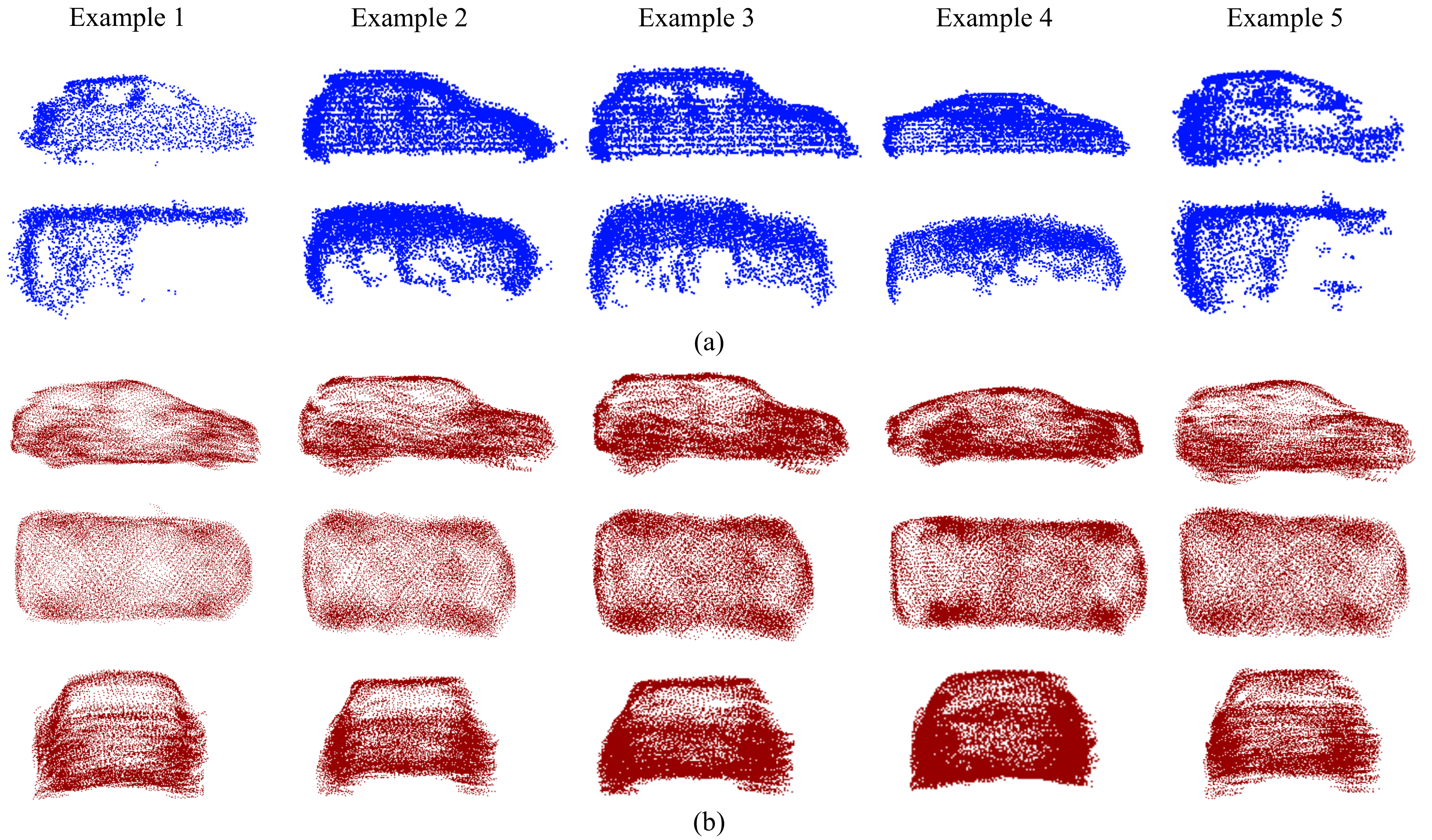}
		\caption{Completed 3D point clouds using real-scan data from the TUM-MLS-2016 dataset. (a) Five vehicle examples of partial point clouds seen from side view and top view. (b) Completed point clouds displayed from different viewpoints: side view, top view, and rear view.}
		\label{fig:tum_data}
    \end{center}
\end{figure*}
To further illustrate our method's effectiveness and generalization ability on real scan data, we selected the TUM-MLS-2016 dataset as a test set. 
We do not have complete point clouds as ground truth for the TUM-MLS-2016 dataset either.  
Therefore, we selected  qualitative results of some vehicle instances and show them here in Fig.~\ref{fig:tum_data}. 
Unlike point clouds from the KITTI dataset, point clouds from the TUM-MLS-2016 dataset are very dense. 
These partial point clouds contain 4200 points on average here. 
In spite of this, our method can still generate detailed information not only in   partial inputs but also for the missing structures. 
For example, in the fourth row of Fig.~\ref{fig:tum_data}, the point cloud completed by our method preserves the shape of the input and reconstructs the wheels and other missing parts. 
This verifies that our approach can transfer easily between the different distributions without any fine-tuning operations, whether partial point clouds are from the KITTI dataset,  the TUM-MLS-2016 dataset, or  the ShapeNet dataset.

\subsubsection{Application}\label{sec:Appilication}
\begin{figure*}[ht]
	\begin{center}
		\includegraphics[width=16cm]{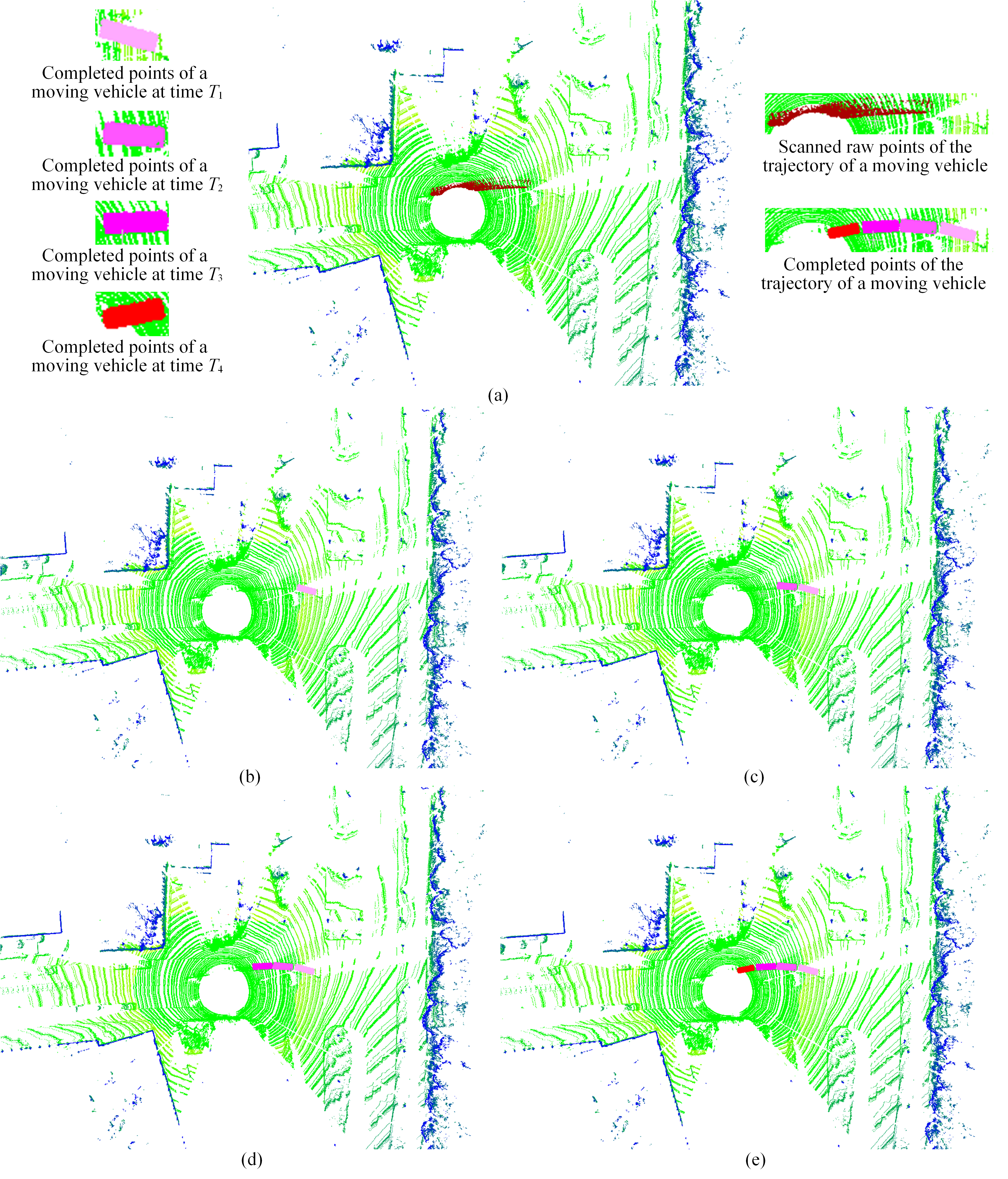}
		\caption{Application to 3D traffic monitoring. (a) 3D traffic scene at the crossroad visualized using  the SLAM technique. (b)-(d) Different colorful point clouds represent the completed point clouds by VPC-Net of this vehicle appears at different times.}
		\label{fig:monitoring}
    \end{center}
\end{figure*}
Apart from evaluating the effectiveness of the proposed method,  more complete and denser point clouds can be helpful  for many common tasks \cite{yuan2018pcn}. 
We   applied the completed results to a 3D vehicle monitoring task.
The proposed method VPC-Net can provide complete shape information about vehicles, which can be regarded as an assistant for this task. It also demonstrates that the proposed method is suitable for real-time applications. Note that we do not handle   existing issues in  the monitoring task using completed vehicles. The goal is to provide the shape of the vehicles for  the monitoring task only based on the existing raw LiDAR data.

Therefore, one Velodyne HDL-64E rotating 3D laser scanner is placed on the center of crossroads to collect the spatially dense and accurate 3D information. 
The round hole in Fig.~\ref{fig:monitoring}a is the location of the LiDAR system. 
The typical monitoring technique Simultaneous Localization and Mapping (SLAM) \cite{cadena2016past} is leveraged to estimate the vehicles in a 3D map while simultaneously localizing the object within it. 
The velocity, orientation, and trajectory of vehicles can be obtained using the SLAM method. 
However, it cannot reconstruct the complete shape of moving vehicles, as shown in Fig.~\ref{fig:monitoring}a. 
 In Fig.~\ref{fig:monitoring}a, the brown point clouds represent a moving car passed in 
  this LiDAR-based system, and form   a band shape. 
For such dynamic vehicles, we detected them from each frame's raw data and completed them by the VPC-Net trained on the ShapeNet dataset. 
Figs.~\ref{fig:monitoring}b-d show the completed vehicle appeared on these crossroads at continuous time  $T_1$,  $T_2$,  $T_3$,  and  $T_4$, respectively. 
As can be seen, the proposed method VPC-Net can be applied to the real-time 3D vehicle monitoring task. 
Furthermore, the completed point clouds have full-content information on vehicle models. As pointed out by \cite{pan2018novel}, the complete shape of the measured vehicles plays an important role in designing the structure of urban highway viaducts, since it is   key to estimating wind pressure caused by   vehicles driving close to the sound barrier. Thus, the complete shape of vehicles will help   traffic managers make the right decisions when designing   highway viaducts. In addition, the 3D shape acquisition of vehicles is   critical   in  the dynamic 3D reconstruction of traffic on road tasks~\cite{zhang2020vehicle}. However, they used the 3D CAD vehicle models from ShapeNet instead of real vehicles to simulate real traffic scenes. This strategy cannot deal with occluded vehicles, nor can it preserve the real shape knowledge of them. From this point of view, the shape of moving vehicles completed by our proposed method can support   dynamic 3D traffic scene reconstruction tasks.

\section{Discussion}\label{sec:Discussion}

\subsection{Visualization of completion details}\label{sec:Visualization of completion details}

\begin{figure*}[!tb]
	\begin{center}
		\includegraphics[width=16cm]{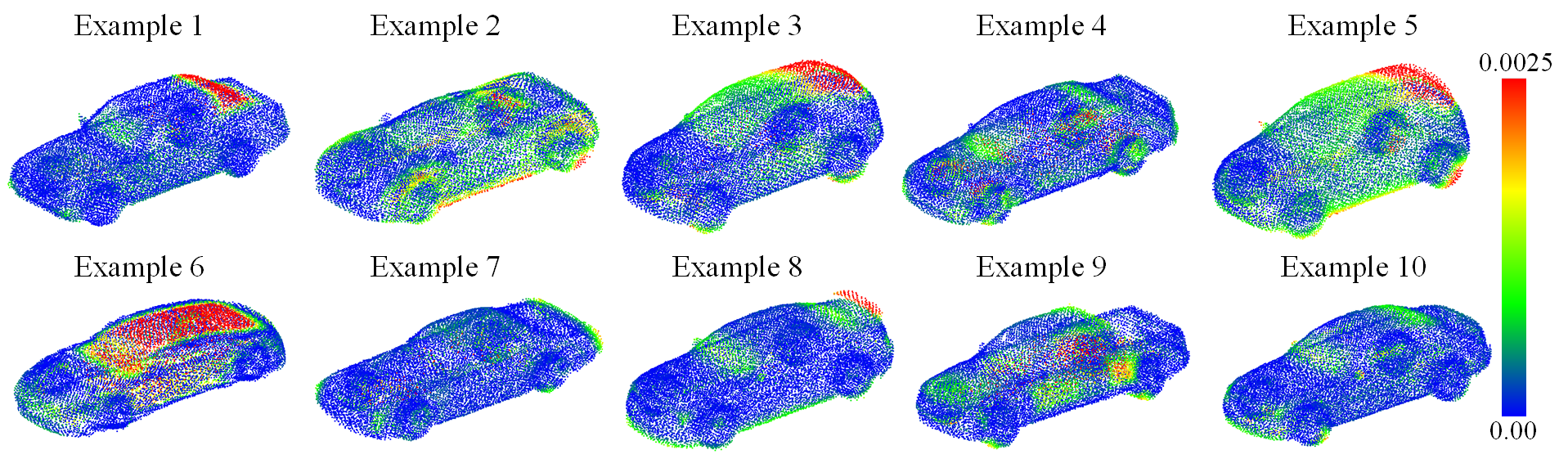}
		\caption{Visualizing point distances between the completed point clouds with ground truth point clouds.}
		\label{fig:DistanceResults}
    \end{center}
\end{figure*}

To better gain further insights about the details of completion performance, we visualize the residual distance between corresponding points from the outputs of our method VPC-Net to the ground truth in Fig.~\ref{fig:DistanceResults}. 
The 10 different vehicles are from ShapeNet test data. 
This figure provides detailed information about which vehicle parts were completed correctly. Different colors encode the normalized distance between the corresponding shapes. 
Fig.~\ref{fig:DistanceResults}  clearly shows that the output point clouds completed by the proposed VPC-Net recovered   most of vehicle parts correctly. 
In addition, by observing the red area in Examples 1, 3, 5, and 6, it can be seen that our method cannot capture the fine-grained details in terms of the roof of vehicles. 
However, from the perspective of human perception, it can be tolerated since humans tend to judge an object's quality by its global features and will tolerate small inaccuracies in shape or location \cite{tatarchenko2019single}.

\subsection{Ablation study}\label{sec:Ablation study}

The ablation studies evaluated the effectiveness of the different proposed components in our network, including  the spatial transform network (STN),  the point feature enhancement operation (PFE), and  the refiner. 
We developed four models: (1) a model without STN, PFE, or  the  refiner, (2) a model with STN only, (3) a model with both STN and PFE, and (4) a model with STN, PFE, and  the refiner. 
We used CD and EMD as the evaluation metric, and the quantitative results of these models are shown in Table \ref{tab:Tab2}.  
All experiments were conducted on the ShapeNet dataset, and the resolution of  the points was 16,384. 
It is clear that our full pipeline has the best performance.

As shown in Table \ref{tab:Tab2}, with the proposed STN module, our model achieves an improvement of 23.5$ \% $ and 14.1$ \% $ on CD and EMD, respectively. 
This is because the rigid geometric transformation has a significant effect on extracting features from partial inputs, while STN can learn invariance to translation and rotation.  
With the proposed PFE module, our model   improves (0.1 $ \% $, 0.3 $ \% $) the CD and EMD.
This confirms that enhancing the global feature is essential to   generate a more accurate coarse point cloud. 
The proposed refiner module can further improve the performance by 3 $ \% $ and 0.2 $ \% $ in terms of the CD and EMD. 
The improvement in the CD is especially significant. 
This is because the refiner actually improves the fine-grained details of  the completed point clouds,  and  the CD is better for measuring the fine-detailed structure of objects than  the EMD. As pointed out in \cite{fan2017point},  the CD will produce   points outside the main body at the correct locations. The EMD roughly captures the mean shape and is considerably distorted, which means it will ignore some flying but correct points.
The ablation studies demonstrate that each proposed module plays significant roles in our network for performance improvements.
Removing any modules will decline the performance, which proves that each proposed module contributes. 

\begin{table}\small
    \centering
    \caption{Performance comparison of the proposed method with different components. The mean Chamfer Distance (CD) and Earth's Mover Distance (EMD) per point are reported, multiplied by $10^{3}$ and $10^{2}$, respectively.}    
        \begin{tabular}{ccc|cc} 
            \toprule
            STN & PFE & Refiner & CD & EMD     \\ 
            \midrule
              &     &                     & 11.668 & 6.0480  \\
             \checkmark   &     &         & 8.922 & 5.1947  \\
             \checkmark   & \checkmark     &         & 8.916 & 5.1777  \\
             \checkmark   & \checkmark    & \checkmark        & \textbf{8.662} & \textbf{5.1677}  \\
            \bottomrule
        \label{tab:Tab2}
        \end{tabular}
\end{table}

\subsection{Robustness test}\label{sec: robustness test}

\begin{figure*}[!tb]
	\begin{center}
		\includegraphics[width=16cm]{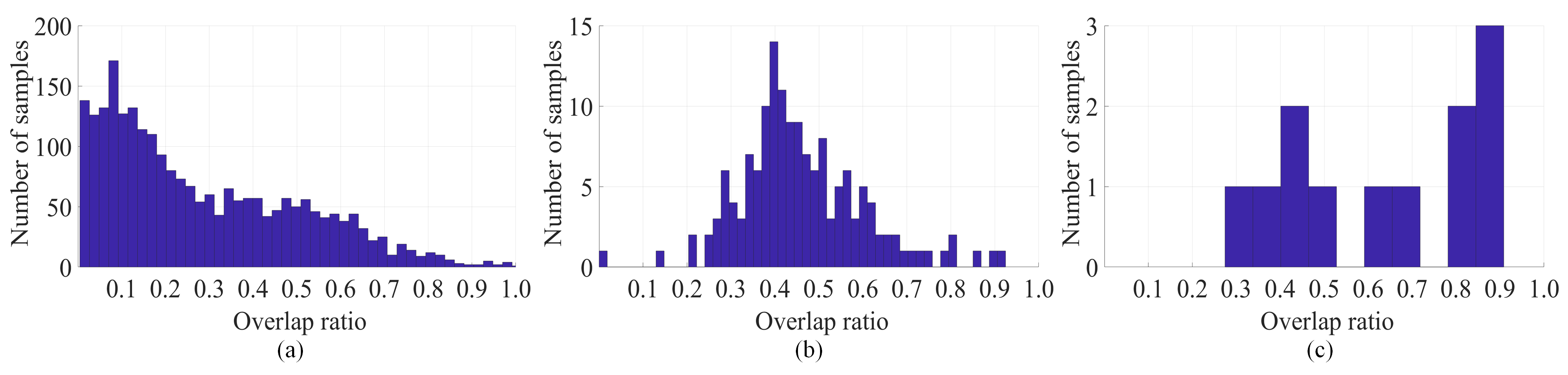}
		\caption{Completeness on the  tested datasets. Overlap ratio between input point clouds and completed point clouds in (a)  the KITTI dataset, (b) the ShapeNet dataset, and (c)  the TUM-MLS-2016 dataset.}
		\label{fig:robust_test_overlap}
    \end{center}
\end{figure*}

\begin{figure*}[!tb]
	\begin{center}
		\includegraphics[width=16cm]{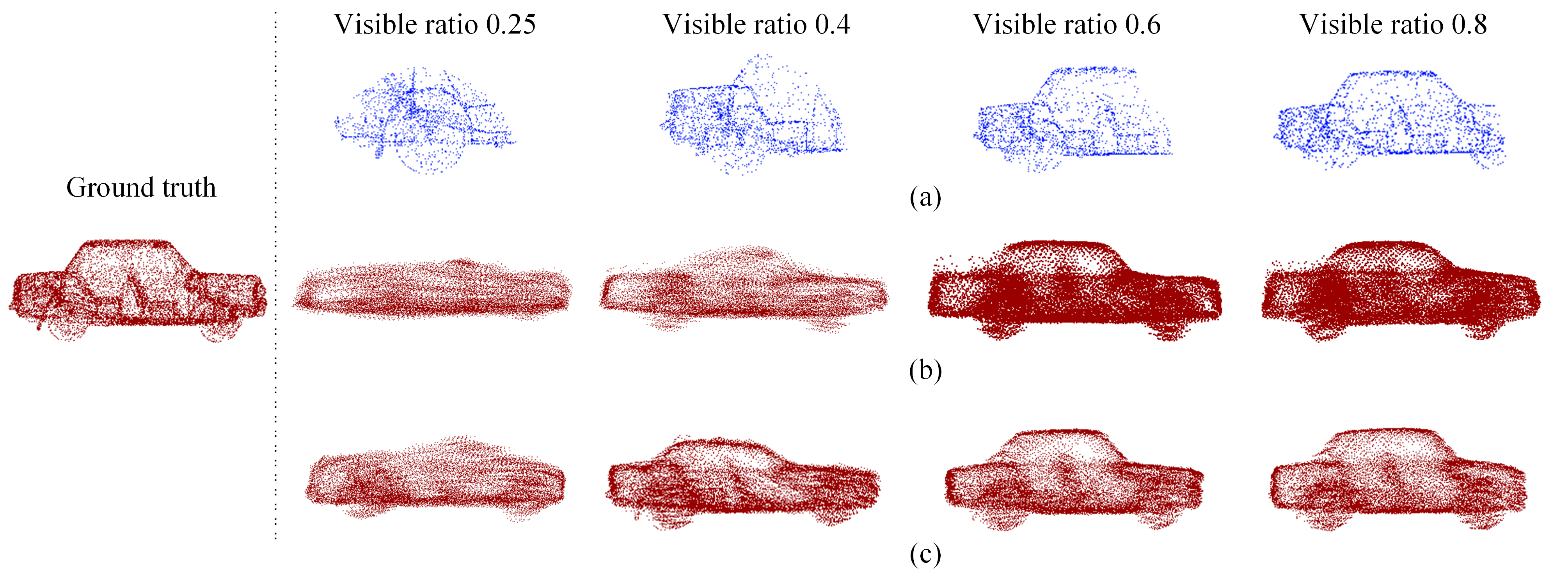}
		\caption{Qualitative results on the  inputs with different amounts of missing content. (a) Partial point clouds with different levels of visibility. Completed point clouds by (b)  PCN and (c)   VPC-Net.}
		\label{fig:robust_test}
    \end{center}
\end{figure*}

\begin{table}\small
    \centering
    \caption{Quantitative results on inputs with different amounts of missing content. The CD is reported by   PCN and our method, multiplied by $10^{3}$.} 
        \begin{tabular}{ccccc} 
            \toprule
            Visible Ratio & 25\%           & 40\%            & 60\%           & 80\%             \\ 
            \midrule
            PCN~\cite{yuan2018pcn}          & 21.555 & 13.979  & 12.002 & 11. 884  \\
            VPC-Net          & \textbf{14.786}  & \textbf{12.377}  & \textbf{7.926} & \textbf{7.612}    \\
            \bottomrule
        \label{tab:Tab3}
        \end{tabular}
\end{table}

We carried out  experiments to evaluate the method's robustness for input point clouds with various   degrees of missing information. 
First, the completeness statistics of the  test data from the ShapeNet dataset,  the KITTI dataset, and  the TUM-MLS-2016 dataset were collected, as shown in Fig.~\ref{fig:robust_test_overlap}. 
We refer to the overlap ratio $R_o$ between the  input partial point clouds and  the completed point clouds as the completeness metric, which is defined by
\begin{equation}
    R_o = S_p / S_c
\end{equation}
\noindent 
where $S_p$ and $S_c$ are surface areas of input partial point clouds and completed point clouds, respectively.

As can be seen in Fig.~\ref{fig:robust_test_overlap}, most   input instances from  the KITTI dataset are very sparse, and   completeness is less than 50$ \% $. 
In contrast, the examples from the TUM-MLS-2016 dataset have enough completeness since that dataset provides the aggregated point clouds, not the original scan data. 
The completeness of  the test data from the ShapeNet dataset is a normal distribution. 
Based on  the experimental results in Section \ref{sec:Experimental results}, our   VPC-Net can handle   these inputs with different completeness. 

To better illustrate the robustness, we performed the robustness test experiment on the ShapeNet test data since there are ground truth point clouds. 
We changed the incompleteness 
  degree $ \textit{d} $ of input point clouds, where $ \textit{d} $ ranges from 20$\%$ to 75$\%$. 
The qualitative and quantitative results are shown in Fig.~\ref{fig:robust_test} and Table \ref{tab:Tab3}, respectively. 
The visible ratios 0.25, 0.4, 0.6, and 0.8 mean that four incomplete inputs lack 75$\%$, 60$\%$, 40$\%$, and 20$\%$ of the ground truth data, respectively. 
As illustrated in Fig.~\ref{fig:robust_test} and Table \ref{tab:Tab3},  we can draw three conclusions: 
(1) Our method is more robust than PCN when dealing with a high   degree of incompleteness. 
For example, when the visible ratio is 0.25, our method is able to generate the general shape of the car, but PCN   fails. 
(2) When more regions are missing, CD and EMD errors slowly increase. 
This implies our method is still robust when meeting   inputs with different incompleteness degrees. 
(3) The outputs completed by both methods are plausible when dealing with incomplete inputs with a large percentage of missing information.
For example, the car generated by our method   is a cabriolet, while the ground truth is a non-convertible car. 
However, this ambiguity is a common issue \cite{fan2017point}, because even for humans, it is difficult  to know what this car is like based on just one wheel.

\subsection{Registration Test}\label{sec: Registration test}

An even density and completeness are key factors for a successful registration between two point clouds \cite{xu2019pairwise}. Correspondingly, the registration result can also reflect the quality (e.g., the evenness of point density or  the completeness of points) of the input point clouds \cite{yuan2018pcn}. 
Here, similar to the test conducted in the work of the baseline method PCN \cite{yuan2018pcn}, we also conducted registration experiments between pairs of vehicle point clouds. Comparing the registration accuracy using incomplete and complete  point clouds demonstrates the feasibility of the proposed vehicle point cloud completion method. 
The vehicle point clouds of adjacent frames in the same Velodyne sequence from the KITTI dataset were chosen as test data. We adopted two types of inputs in the registration method: one represents  the partial point clouds from  the real-scan data, while the other represents the  completed point clouds by the proposed VPC-Net. 

\begin{table}[h]\small
	\centering
		\caption{Averaged rotation and translation errors of point cloud registration using different inputs.}
		\begin{tabular}{l c c}
		    \toprule
			\multirow{2}{*}{Inputs}  &\multicolumn{2}{c}{Average error}\\
			\cline{2-3}
			                 & rotation ($^{\circ}$)    &   translation (m) \\
			\hline
			Partial inputs  & 13.9422                  &   7.0653          \\
			Complete inputs & \textbf{7.9599}          &   \textbf{4.2059} \\
            \bottomrule
		\end{tabular}
    \label{tab:point cloud registration}
\end{table}
\begin{table*}[t]\small
    \centering
    \caption{Quantitative comparison of point cloud registration task with different inputs.}
        \begin{tabular}{c cc cc} 
        \toprule
        \multirow{2}{*}{\begin{tabular}[c]{@{}c@{}}\\ \end{tabular}} & \multicolumn{2}{c}{Partial inputs} & \multicolumn{2}{c}{Complete outputs}  \\ 
            \cline{2-5}
            Example                                                             & Rotation error  & Translation error & Rotation error  & Translation error  \\ 
            \hline
            1                                                    & 4.5159          & 1.4715            & \textbf{1.8219} & \textbf{0.5904}    \\
            2                                                    & 11.4627         & 2.1093            & \textbf{0.5678} & \textbf{0.1060}    \\
            3                                                    & 4.5159          & 1.4715            & \textbf{1.8219} & \textbf{0.5904}    \\
            4                                                    & 143.9396        & 58.0907           & \textbf{1.5606} & \textbf{0.7201}    \\
            5                                                    & 178.6335        & 54.5161           & \textbf{3.1471} & \textbf{1.5235}    \\
            6                                                    & 14.8757         & 7.8894            & \textbf{2.4544} & \textbf{1.2499}    \\
            7                                                    & 3.1952          & 1.8321            & \textbf{2.4083} & \textbf{1.3489}    \\
            8                                                    & 1.7482          & 0.6973            & \textbf{0.9957} & \textbf{0.2084}    \\
            9                                                    & \textbf{0.0270} & \textbf{1.3128}   & 5.5954          & 3.0927             \\
            10                                                   & \textbf{0.6646} & \textbf{1.3941}   & 4.1969          & 3.8149             \\
            \bottomrule
        \end{tabular}
    \label{tab:point cloud registration examples}
\end{table*}
\begin{figure*}[!tb]
	\begin{center}
		\includegraphics[width=16cm]{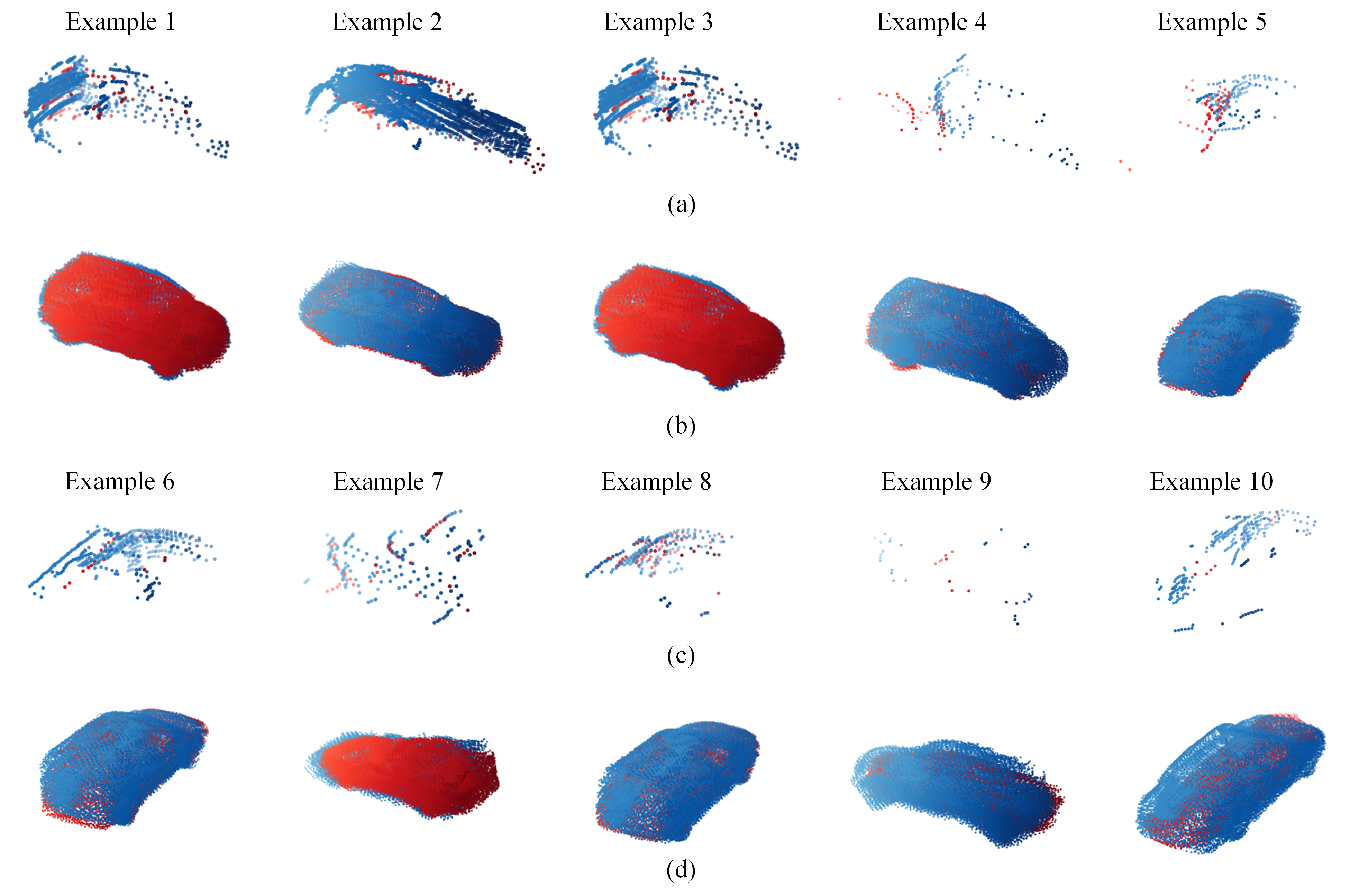}
		\caption{Qualitative comparison of point cloud registration task with different inputs.  (a) and (c) Registered results with partial point clouds. (b) and (d) Registered completed results of the same examples.}
		\label{fig:kitti_registration}
    \end{center}
\end{figure*}
Moreover, a simple point-to-point ICP \cite{besl1992method} was applied as a registration algorithm, which minimizes distances iteratively between points from two point clouds. 
Notably, the ICP algorithm is not the only choice for registration tasks. Any registration algorithm that can be applied to illustrate the completed results has a good and consistent shape for the same vehicle instances in different frames.
The average rotational and translational error in the registration results with partial and complete input point clouds were compared. 
The rotational error $E_R$ and translational error $E_T$ are defined as follows, respectively:
\begin{equation}
    E_R=2\cos ^{-1}(2< R_1,R_2>^{2}-1 )
\end{equation}
\begin{equation}
    E_T=||T_1-T_2||_2
\end{equation}
\noindent where $R_1$ and $T_1$ are the rotation and  translation of the  ground truth in the KITTI dataset, respectively. $R_2$ and $T_2$ are the rotation and translation measured by the ICP method, respectively.

As shown in Table \ref{tab:point cloud registration},  the quantitative results demonstrate that   the complete point clouds generated by VPC-Net provide  a more accurate estimation of translation and rotation than   the    incomplete point clouds when   conducting the registration test. 
Specifically,  rotation and translation accuracy   improves by 42.9$\%$ and 40.5$\%$, respectively. 
In Fig.~\ref{fig:kitti_registration}, 10 qualitative examples are displayed. The completed point clouds have large overlapping regions recovered by VPC-Net, which demonstrates that   VPC-Net can generate consistent shapes with high quality for the same vehicle in different frames. 
We list the corresponding rotation and translation errors for these examples in Table \ref{tab:point cloud registration examples}. 
As can be seen from Example 1 to Example 8, the registration using complete point clouds shows an improvement in both rotation and translation accuracies. 
The improvement is most significant when the error with partial inputs is relatively large.
Examples 9 and 10 are   failure cases where 
  the registered partial inputs have better performance than registered complete inputs. 
However, this is explained by   the qualitative results in Fig.~\ref{fig:kitti_registration}:   the registered partial inputs have too few points, only  about 10, so   the ICP method is not able to compute the errors accurately. 

\section{Conclusion}\label{sec:Conclusions}

In this paper, we propose a novel end-to-end network, VPC-Net, for vehicle point completion using sparse and partial point clouds. Our method can generate complete and realistic structures  and can maintain   fine-grained details in an efficient manner. Furthermore, it is effective across different resolutions of inputs. Experimental results on the ShapeNet dataset,  the KITTI dataset, and  the TUM-MLS-2016 dataset demonstrate the effectiveness of our proposed VPC-Net compared to    state-of-the-art methods. It also has a strong generalization performance on real-scan datasets, which makes it   suitable and beneficial for practical applications. The main benefits of the VPC-Net can be summarized as follows:
\begin{itemize}
\item The quality of  the point clouds has less influence on   VPC-Net, which indicates that the proposed method is more robust given  various resolutions and   varying degrees of missing  point clouds compared to other 3D point cloud completion networks.
\item   VPC-Net provides satisfying results for various datasets, which is attributed 
  to three aspects. The first one is the PFE layer, which combines   low-level local features and high-level semantic features. Second, the spatial transformer network guarantees that extracted features are invariant to rigid rotation and translation. The third one is the refiner module, which tends to preserve the fine details of input point clouds.
\end{itemize}
However, there are   limitations of the proposed method. For example, the designed refiner will increase the number of training parameters compared with  the previous point cloud completion network PCN. Considering the ambiguity of the completion at test time, in the future, we will   generate multiple plausible shapes and then assess the plausibility of several various completions. 
We  will also plan to complete other objects in   urban scenes, such as buildings, traffic signs, road lanes, and so on.

\section*{Acknowledgments}\label{sec:Acknowledgments}
This research was funded by  the Natural Science Foundation of China, grant number U1605254. This work was carried out within the frame of Leonhard Obermeyer Center (LOC) at Technische Universität München (TUM) [www.loc.tum.de]. We would like to thank Weiquan Liu and Qing Li for their suggestions and support. This research was supported by the China Scholarship Council.
%
%


\bibliographystyle{ieee}
\bibliography{egpaper_final.bbl}
\end{document}